\documentclass[utf8]{frontiersSCNS} 
\usepackage{url,lineno,microtype,subcaption}
\usepackage[onehalfspacing]{setspace}

\usepackage[english]{babel}

\usepackage{paralist} 
\usepackage{siunitx} 
\usepackage{multirow} 
\usepackage[hidelinks]{hyperref}
\usepackage[capitalize]{cleveref} 
\graphicspath{ {Figures/} }
\usepackage{tabularx} 
\usepackage{rotating} 

\usepackage{tikz}
\usepackage{changes}

\DeclareRobustCommand{\inlinelist}[1]{\begin{inparaenum}[(a)] #1 \end{inparaenum}}

\newcommand{\mps}{\meter\per\second}
\newcommand{\jnm}{J\per (Nm)}
\newcommand{\us}{\micro\second}
\newcommand{\ms}{\milli\second}

\newcommand*\circled[1]{\tikz[baseline=(char.base)]{
            \node[shape=circle,draw,inner sep=1pt] (char) {#1};}}

\def\keyFont{\fontsize{8}{11}\helveticabold }
\def\firstAuthorLast{Spr\"owitz {et~al.}} 
\def\Authors{
Alexander Spr\"owitz\,$^{1,2,*}$, 
Alexandre Tuleu\,$^{1}$, 
Mostafa Ajallooeian\,$^{1}$, 
Massimo Vespignani\,$^{1}$, 
Rico M\"ockel\,$^{1,3}$, 
Peter Eckert\,$^{1}$, 
Michiel D'Haene\,$^{4}$, 
Jonas Degrave\,$^{4}$, 
Arne Nordmann\,$^{5}$, 
Benjamin Schrauwen\,$^{4}$, 
Jochen Steil\,$^{5}$, 
and Auke Jan Ijspeert\,$^{1}$}

\def\Address{
$^{1}$Biorobotics Laboratory, Institute of Bioengineering, \'Ecole polytechnique f\'ed\'erale de Lausanne (EPFL), Lausanne, Switzerland \\
$^{2}$Dynamic Locomotion Group, Max Planck Institute for Intelligent Systems, Stuttgart, Germany \\
$^{3}$Department of Knowledge Engineering, Maastricht University, The Netherlands \\
$^{4}$Department of Electronics and Information Systems, Ghent University, Gent, Belgium \\
$^{5}$CoR-Lab, Research Institute for Cognition and Robotics, Bielefeld University, Germany}

\def\corrAuthor{Heisenbergstr. 3, 70569 Stuttgart}

\def\corrEmail{sprowitz@is.mpg.de}

\begin{document}
\onecolumn
\firstpage{1}

\title[Oncilla robot]{Oncilla robot: a versatile open-source quadruped research robot with compliant pantograph legs}
\author[\firstAuthorLast ]{\Authors} 
\address{} 
\correspondance{} 
\extraAuth{}

\maketitle
\begin{abstract}
We present Oncilla robot, a novel mobile, quadruped legged locomotion machine. 
This large-cat sized, \SI{5.1}{\kg} robot is one of a kind of a recent, bioinspired legged robot class designed with the capability of model-free locomotion control.
Animal legged locomotion in rough terrain is clearly shaped by sensor feedback systems. 
Results with Oncilla robot show that agile and versatile locomotion is possible without sensory signals to some extend, and tracking becomes robust when feedback control is added~\citep{mos-thesis}.
By incorporating mechanical and control blueprints inspired from animals, and by observing the resulting robot locomotion characteristics, we aim to understand the contribution of individual components.
Legged robots have a wide mechanical and control design parameter space, and a unique potential as research tools to investigate principles of biomechanics and legged locomotion control.
But the hardware and controller design can be a steep initial hurdle for academic research.
To facilitate the easy start and development of legged robots, Oncilla-robot's blueprints are available through open-source.

The robot's locomotion capabilities are shown in several scenarios.
Specifically, its spring-loaded pantographic leg design compensates for overdetermined body and leg postures, i.e.\,during turning maneuvers, locomotion outdoors, or while going up and down slopes. 
The robot's active degree of freedom allow tight and swift direction changes, and turns on the spot. 
Presented hardware experiments are conducted in an open-loop manner, with little control and computational effort.
For more versatile locomotion control, Oncilla-robot can sense leg joint rotations, and leg-trunk forces. 
Additional sensors can be included for feedback control with an open communication protocol interface.
The robot's customized actuators are designed for robust actuation, and efficient locomotion. 
It trots with a cost of transport of \SI{3.2}{\jnm}, at a speed of \SI{0.63}{\mps} (Froude number \num{0.25}).
The robot trots inclined slopes up to \SI{10}{\degree}, at \SI{0.25}{\mps}.
The multi-body Webots model of Oncilla robot, and Oncilla robot's extensive software architecture enables users to design and test scenarios in simulation. Controllers can directly be transferred to the real robot.
Oncilla robot's blueprints are open-source published (hardware GLP v3, software LGPL v3).

\tiny \keyFont{ \section{Keywords:} quadruped, robot, pantograph, open-source, multiple gaits, open-loop, pattern generator, turning}
\end{abstract}
 \section{Introduction}
Emerging technologies allow to improve our understanding of legged locomotion, and its underlying principles.
We argue that custom-designed, bioinspired legged machines like the presented Oncilla robot have the potential to provide valuable insights into biomechanics and neuromuscular control of animal legged locomotion. 
We show initial robot locomotion examples in simulation and hardware, directed with minimal control effort.

Technological progress shapes research, for instance traditional dissection tools are being expanded by computer guided imaging. 
3D scans provide spatial data of the animal's morphology even in motion~\citep{walker_vivo_2014}.
Advanced computer simulations give insights into complex interactions between muscles, tendons, and skeletal structures during locomotion~\citep{delp_opensim:_2007}.
This allows estimating movements, forces and interactions of otherwise hidden and unaccessible structures in animals.
With the help of new technologies, it is the goal to find necessary and sufficient biomechanical and neuromuscular control components for legged locomotion, by identifying form and function.

However, mapping morphologies to function is not trivial. Evolutionary byproducts such as spandrels can mask function of structures~\citep{gould_spandrels_1979}.
The animal legged apparatus developed for locomotion---a complex task with hybrid dynamics. Balancing, carrying and accelerating the animal's body requires precise movement coordination through feedback, and redirection of loads. Forces and movements largely differ in swing and stance phase, separated by harsh impacts at landing, and high velocities during toe off~\citep{sohnela_kinetics_2017}.

Our approach includes custom-made, bioinspired legged robot hardware, and its locomotion controller. Oncilla robot allows us to tackle locomotion research from a new perspective, and directly evaluate functional morphologies.
With robotic setups we have the freedom to implement and test simplified `blueprints', mimicking aspects of animals' biomechanics and neuromuscular control. We understand such `blueprints' as building plans which are transferred from Biology to Robotics, for the purpose of testing and characterizing them. The pantographic leg configuration described by \citet{witte_quadrupedal_2003} is an example for a mechanical blueprint, that was transferred from animal biomechanics. It is based on the observation of the mostly parallel orientation of distal and proximal limb segments in three-segmented mammalian legs, during locomotion. Mathematical models of the spinal cord activation i.e. in lampreys, so called 'Central Pattern Generators', are an example for a neuromuscular control blueprint. They were hypothesized and observed in animals, and tested and implemented in robots~\citep{ijspeert_swimming_2007}.

Robotic implementations can swiftly be altered, i.e.\,to test effects of scaling, while longitudinal animal studies are significantly more time consuming.
Testing features in robot hardware allows one to quantify function; components' states (position, velocity etc.), forces linking components, and internal loads can be measured with dedicated sensors. 
Signals for locomotion control and sensing are observable, and mostly controlled. In contrast, recording of deep tissue movements and displacements, and their interacting forces in animals is often not feasible. 
At present recording of animal nerve signals is limited to simpler setups~\citep{daley_role_2009}.

Computer simulated models of legged animals and legged robots forcibly introduce simplifications, compared to real world physics.
Several aspects of physics are notably hard to simulate both precisely and efficiently: leg-ground impacts, wobbling masses, damping, compliance and friction~\citep{schmitt_human_2011}.
Correctly simulating a legged animal or robot walking through soft substrate is difficult~\citep{falkingham_birth_2014,li_sensitive_2009}. 
In comparison, robotic hardware can be a realistic physical (as opposed to numerical) model of an animal---legs and feet are physically interacting with locomotion substrates through contact forces. 
Still, computer models are immensely insightful, also because they are relatively easy to derive and apply. 
Here, we derived a computer simulated Oncilla robot model to rapidly test controller parameters, and create  locomotion data.

A future iterative robotic approach to legged research can be composed of four steps:
a) implementing blueprints of mechanics and control of legged animals, in a legged robot system,
b) creating locomotion case scenarios with legged robots,
c) collecting and analyzing locomotion data, and eventually
d) implementing alterations to initial blueprints. 
If successful, improved blueprints will decrease the discrepancy between robot and animal locomotion characteristics.
Changes and observed invariants allow insights into underlying aspects of legged locomotion. 

From a design and testing perspective, the cyclic iteration works best if tested blueprints can be altered swiftly. Oncilla robot is designed modularly, i.e. its legs and trunk, actuators, sensors, and controllers are relatively easy to replace.
Research in legged locomotion progresses iteratively, and fewer researchers are interested in designing entire robotic systems.
The Oncilla robot open source project allows researchers to i.e.\;test sub-functions of a new locomotion controller.
Reproducing parts requires a low level of technology; access to a generic 3D printer, and a basic machining workshop. Components such as brushless motors are off-the-shelve items. 
Printed circuit board layouts and firmware code are included in the project source code. 
If required, they can be replaced by custom solutions---Oncilla robot's custom interfaces are open sourced.

A novice user requires a legged robot that is easy to start up and run. 
A robot with a large basin of robust locomotion patterns is therefore beneficial. Ideally, the robot's locomotion characteristic would be sufficiently 'docile'. 
At the same time, the robot should have the potential for complex gait scenarios, such as rapid acceleration, turning, and locomotion on slopes.

Oncilla robot features a spring-loaded leg design enabling simplified locomotion control.
As a consequence, this class of legged robots including 'Bow-leg robot', 'Cheetah-cub robot', Bobcat-robot', and `Ken'~\citep{sproewitz_ijrr,zeglin_bow_1999,khoramshahi2013benefits,narioka2012development} can run and hop with feed-forward (FFW) control and does not require model-based control or feedback mechanisms, for basic locomotion on flat terrain.
These robots run in FFW mode at comparatively low control frequencies:~\cite{sproewitz_ijrr} report RC servo motor control with as little as $f=\SI{50}{Hz}$ control frequency, also in case of step-down perturbations during higher speed running.

In comparison, robots with leg designs like MIT Cheetah~\citep{seok_design_2013} or ANYmal~\citep{hutter_anymal-highly_2016} actively extend leg joints against gravity. Actively extending legs require dedicated controllers for leg length force control~\citep{hyun_high_2014}.

With Oncilla robot we extend the capabilities of its robot class with its passively, spring extending legs; the robot turns rapidly, with minimal turning radii, walks slopes up and down, and outdoors on natural substrates fully mobile, with low computational effort and without the need for sensory feedback.

The remaining paper is organized as follows. 
In the next section, we briefly present related work. \Cref{sec:Mechanical-Hardware} describes the robot's mechanical hardware. 
\cref{sec:sw-arch} explains the software architecture of Oncilla robot. 
Results from hardware and simulation experiments are shown in \cref{sec:experiments}.
We discuss design, results and future directions and provide links to the open-source files in \cref{sec:discussion}, and conclude in \cref{sec:conclusion}. 
The supplementary material provides further details on robot design, kinematics, and experiments.

\section{Mechanical Hardware}\label{sec:Mechanical-Hardware}
This section explains Oncilla robot's design, and its actuator design process.
Details are provided for its leg design, the active degree of freedom (DOF) for steering, force sensors, chassis design, and electronic circuits.

\subsection{Physical robot dimensions}
\begin{figure}[h!]
\begin{minipage}[b]{.3\linewidth}
\centering\includegraphics[width=.98\textwidth]{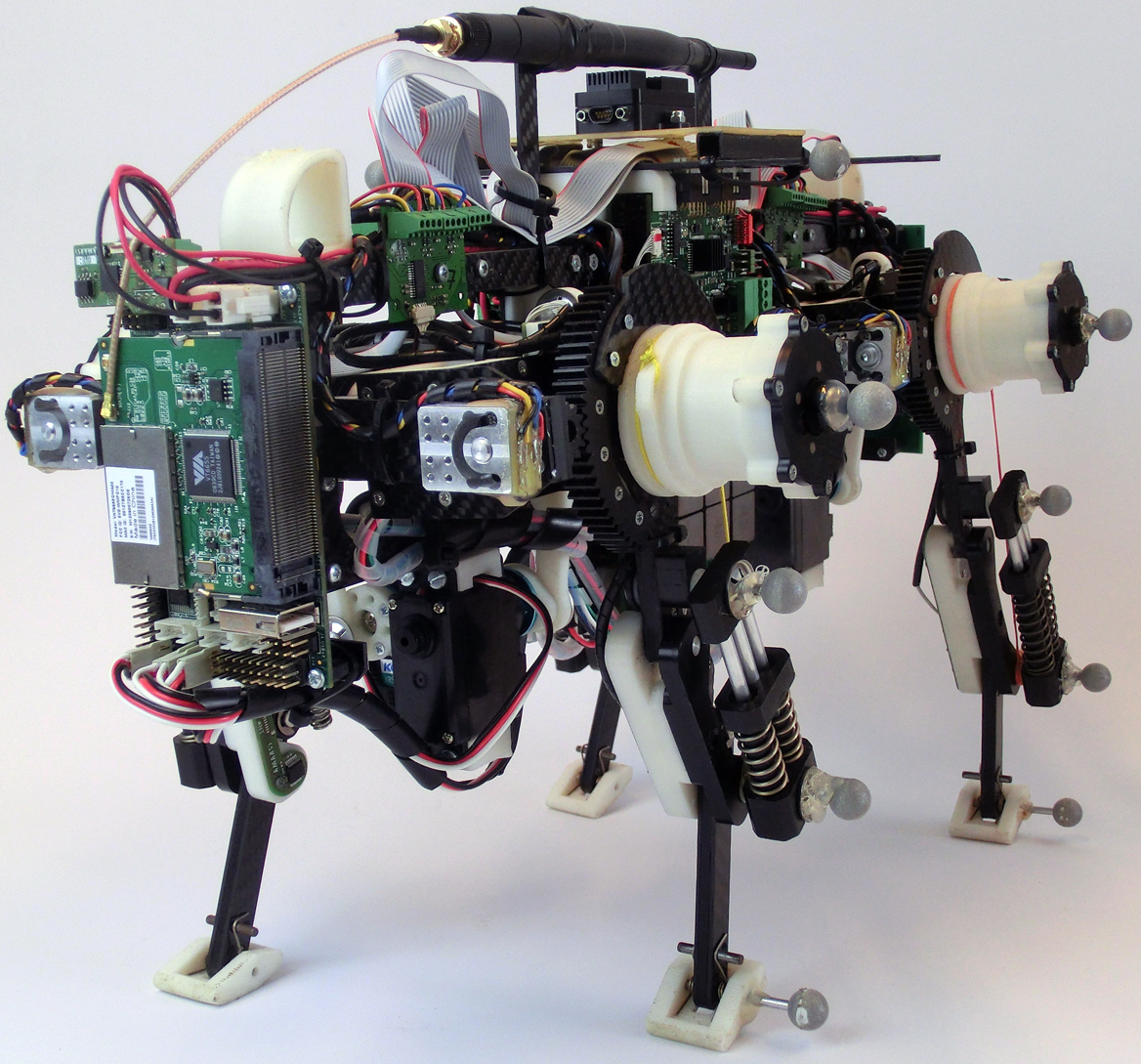}
\end{minipage}
\begin{minipage}[b]{.7\linewidth}
\centering\includegraphics[width=.98\textwidth]{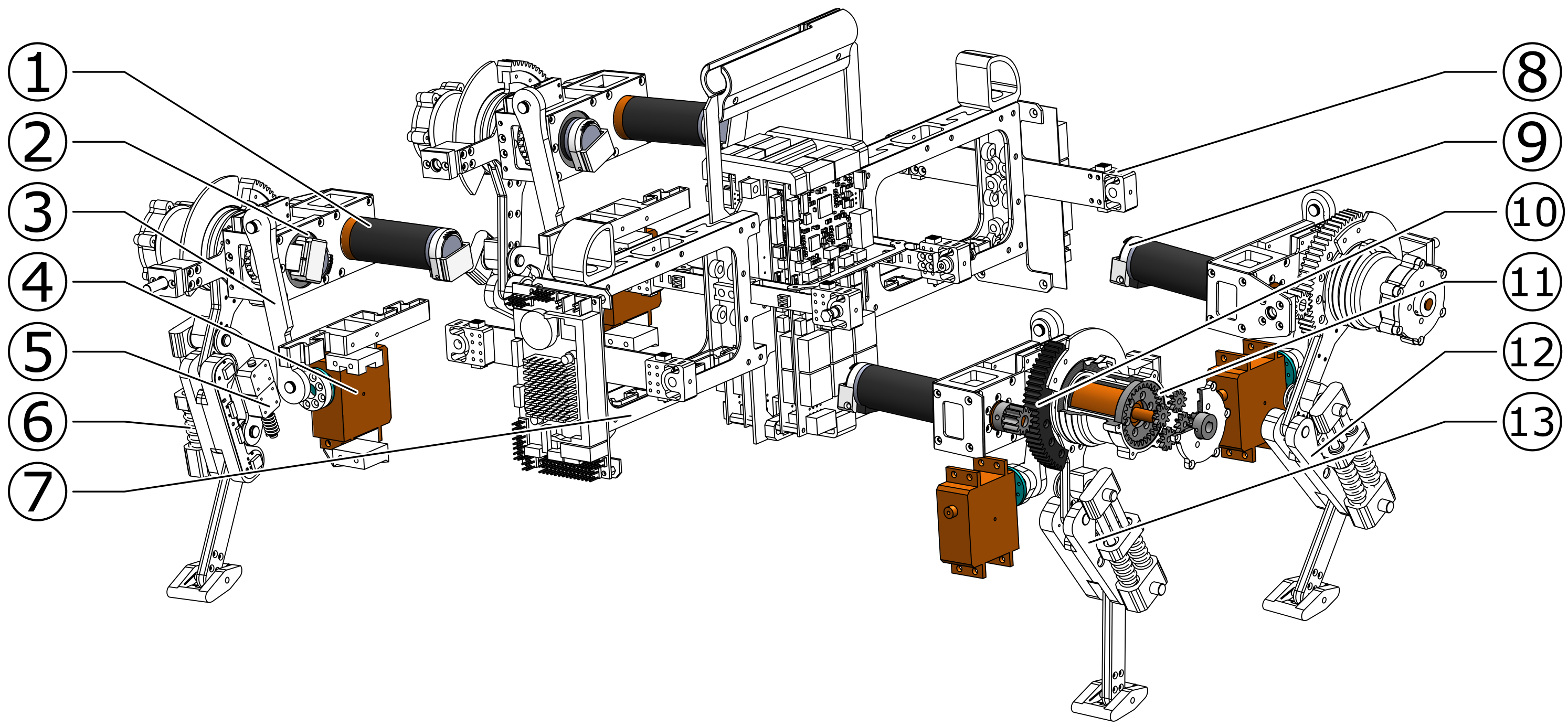}
\end{minipage}
\caption{
Oncilla quadruped robot: \inlinelist{\protect\item photo in isometric view \protect\item exploded view with actuator and mechanical components.}
Exploded view of Oncilla robot, with actuation related components emphasized by coloration. 
The brushless leg angle (LA) motors, are placed parallel and off the hip axis (1),
the brushless leg length (LL) motors are aligned coaxially with the hip axis (2).
A four bar mechanism (3) adducts and abducts (AA joint) the leg,
and is actuated by a RC servo motor (4). 
The parallel leg spring (5) allows the pantograph leg to rotate its distal leg joint under load, 
the leg's diagonal spring (6) acts as the gravity compensating spring. 
Robot trunk is (7).
(8) shows the vertical force sensor of the robot, 
(9) indicates the incremental encoder, mounted at the rear end of each brushless motor.
LA motor actuates the leg through a spur gear pairing, gear ratio $84:1$ (10),
LL motor is geared down with a custom planetary gearbox, gear ratio $56:1$ (11).
(12) shows the hind leg, and (13) the front leg, each pointer indicating the leg's $l_2$ segment.}
\label{fig:oncilla-photo-cad}
\end{figure}

Oncilla robot has the size and weight of a large house cat (\cref{tab:physical-parameters}). It is \SI{0.4}{m} long in total, with \SI{0.23}{m} between front and hind legs, has a maximal hip height of \SI{0.18}{m} with its legs hanging in-air, a hip standing height of \SI{0.16}{m}, and an overall height of \SI{0.31}{m}, from bottom to handle (\cref{fig:frontal-lateral,fig:oncilla-photo-cad}). The robot has \SI{0.14}{m} of lateral spacing between left and right adduction/abduction axes, and a maximum width of \SI{0.25}{m}. Total weight is \SI{4.5}{kg} without, and \SI{5.1}{kg} with a \SI{4500}{mAh} lithium polymer battery pack. The trunk was designed with a center of mass (COM) \SI{36}{mm} below the hip-shoulder axis. Each leg weighs \SI{0.2}{kg} without, and \SI{0.6}{kg} including leg angle and leg length actuators.

\subsection{Actuator design}\label{sec:load-scenario}
Choosing a robust actuator consisting of motor and gearbox is not trivial, when designing robots for agile legged locomotion.
Here we only considered electromagnetic motors, for their ease of use in laboratories.

Mobile robots carry their own weight. This includes the power source, i.e. in form of batteries, but also printed circuit boards (PCB), actuators, chassis, and sensors. 
To reduce power consumption and to increase agility and load carrying capability, low weight and efficient actuators are beneficial.
At least two DOF per leg are required for versatile legged locomotion. 
The leg length (LL) DOF inserts forces to maintain vertical posture. This LL DOF is also used for leg flexion during swing phase.
The sagittal leg angle (LA) DOF supports movements and torques in fore-aft direction, and retracts and protracts the leg.
Legged robot locomotion is a harsh application for electrical motors. 
Oscillating locomotion patterns at high cycle frequencies lead to high velocities in both directions and at high peak torques. 
Those present `external loads'.
Motor and gearbox combinations lead to additional 'internal loads', and can largely exceed external, required loads~\citep{roos_optimal_2006}.
This effect showed strongly with Cheetah-cub robot, a small quadruped robot actuated by high-geared RC servo motors. 
RC servo motors would work well for \SI{1}{Hz} motion frequencies. 
However at \SI{4}{Hz} motion frequencies and larger amplitudes, the over $\num{300}:\num{1}$ gearbox ratio led to overheating problems. 
Only brief experiments with cool-down phases were possible in a repeatable manner~\citep{sproewitz_ijrr}.

To improve actuation compared to Cheetah-cub robot, we simulated external load case scenarios for LL and LA actuator, based on a simplified foot locus. 
We assumed cycle frequencies between \SI{0.5}{Hz} and \SI{3.5}{Hz}, calculated for a trotting quadruped with Oncilla robot's dimensions. 
This external load case data was piped through an actuator optimization framework~\citep{roos_optimal_2006}, creating a simplified-load dynamical motor model.
From this, we chose gearbox ratios in LL ($56:1$) and LA ($84:1$) direction, for a given brushless motor. 
Details for the external and internal load case simulation setup are provided in supplementary documentation.
With an estimated \SI{80}{\%} effective stride length, the modeled robot's cost of transport was calculated~\citep{tucker_energetic_1970}. Results are shown in \cref{fig:COT-3cases}.
The model predicts an asymptotically declining COT, with \SI{7.8}{\jnm} at a low speed of \SI{0.05}{\mps}, a lowest COT of \SI{2.4}{\jnm} at \SI{0.41}{\mps}, and a slightly increasing COT at higher speeds (\SI{2.7}{\jnm} at \SI{0.71}{\mps}). 
We later recorded the hardware robot's instantaneous power during locomotion, which shows a good match.
While underestimating the hardware COT by about \SI{25}{\%}, the simplified-load model did qualitatively predict the asymptotically declining COT characteristics of the hardware robot.

\subsection{Leg and foot design}
\begin{figure}[h!]
\begin{center}
\includegraphics[width=10cm]{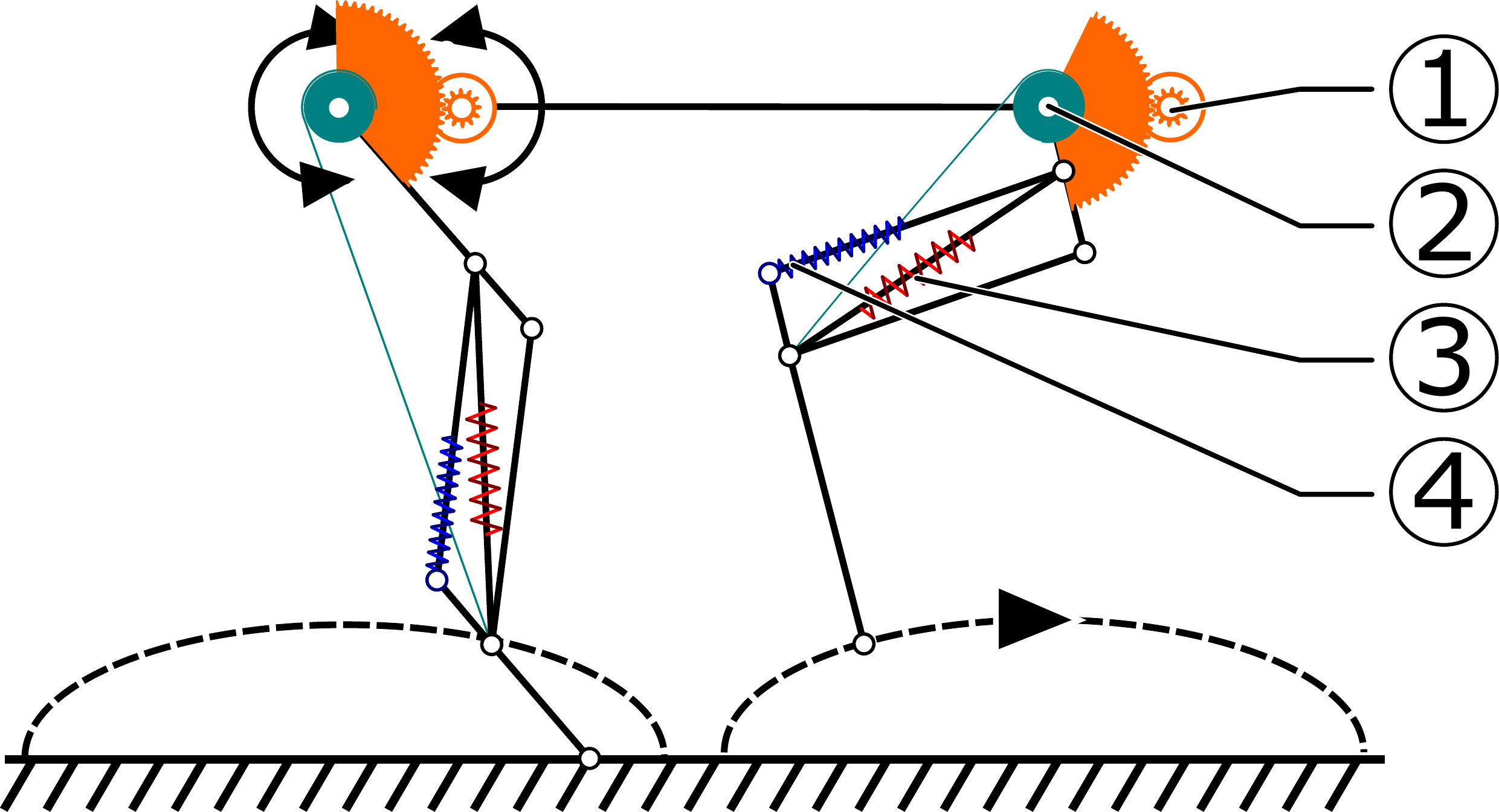}
\end{center}
\caption{
Schematic presentation of Oncilla robot's foot locus movement, created for the simplified-load dynamic-motor (SLDM) model scenario. 
The SLDM model was applied in the robot's pre-design phase, to estimate required motor and gearbox characteristics. 
This foot-locus profile was used to calculate leg length (LL, 2) and (LA, 1) loads, for trot gait. 
The diagonal, gravity compensating leg spring (red, 3) is compressed by flexing the leg through a cable mechanism. 
Load dependent displacement of the parallel spring (4) during stance phase was ignored in the SLDM model. 
}
\label{fig:load-scenario}
\end{figure}

Oncilla robot's leg design is a continuation of Cheetah-cub robot's.
One active DOF was added for adduction/abduction ('AA joint'), to give the robot steering capabilities. 
A set of springs with stiffness \SI{5.8}{N/mm} extends the pantograph leg (red 'diagonal spring', \cref{fig:load-scenario}). One pantograph-segment is replaced with a tensile spring (blue 'parallel spring', stiffness $k=\SI{7.4}{N/mm}$). This spring allows flexion of the distal leg segment under load. The leg extends by \SI{7}{cm}, from its shortest length i.e. during mid-swing, to a fully extended leg length of \SI{18}{cm} in air. A short foot is mounted, spring tensioned by a torsion spring ($k=\SI{1.21}{Nmm\per\degree}$).
Absolute position encoders are placed at hip axes, and two leg joints (supplementary material, Fig.\;S7, $q_0, q_1, q_2$). 
The difference between angles $q_1$ and $q_2$ indicates the external load torque acting at $q_2$, i.e. this joint sensor also acts as analog contact sensor. 
\subsection{Leg length and leg angle actuation}
\begin{figure}[h!]
\begin{minipage}[b]{.5\linewidth}
\centering\includegraphics[width=.98\textwidth]{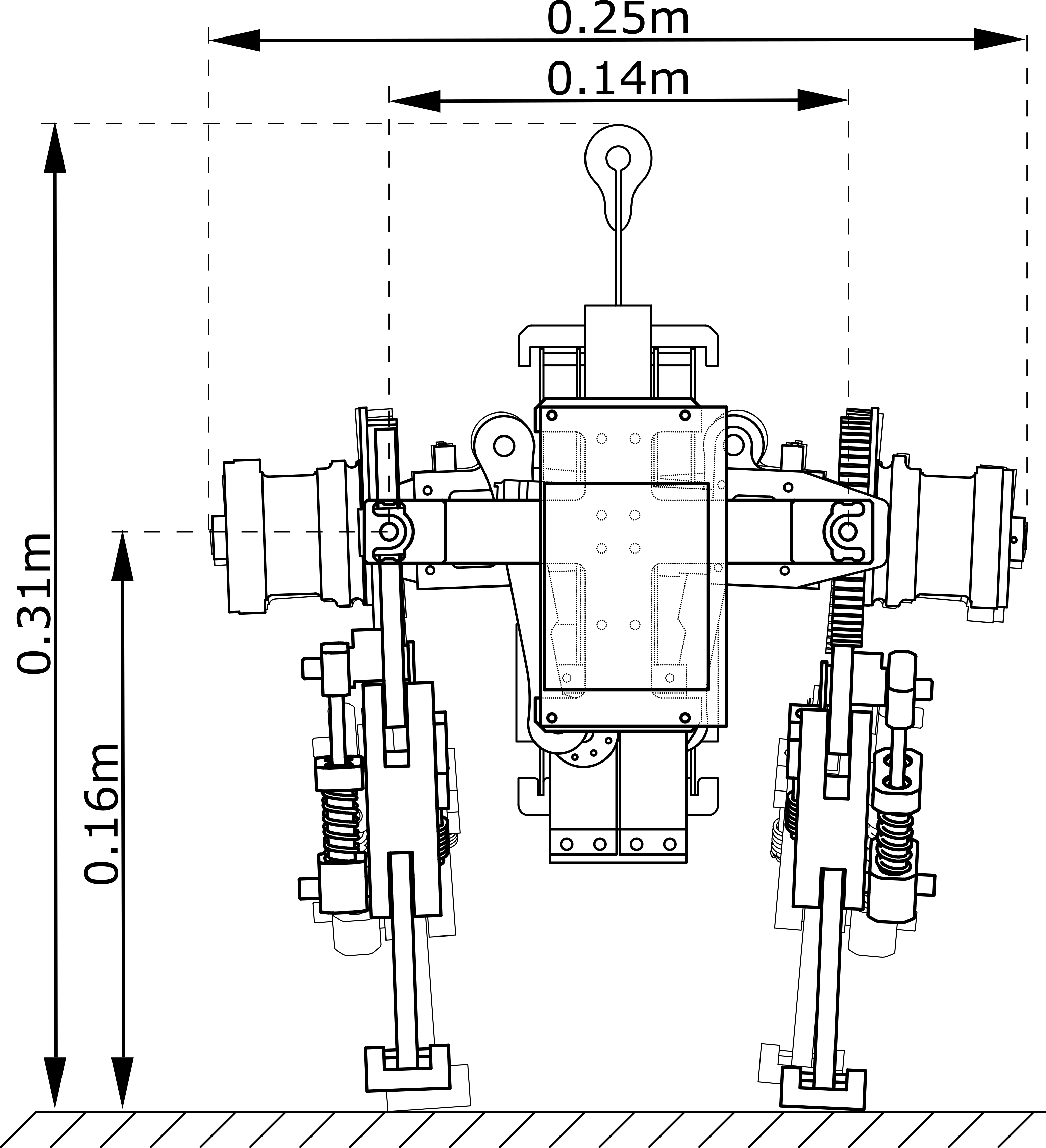}
\end{minipage}
\begin{minipage}[b]{.5\linewidth}
\centering\includegraphics[width=.98\textwidth]{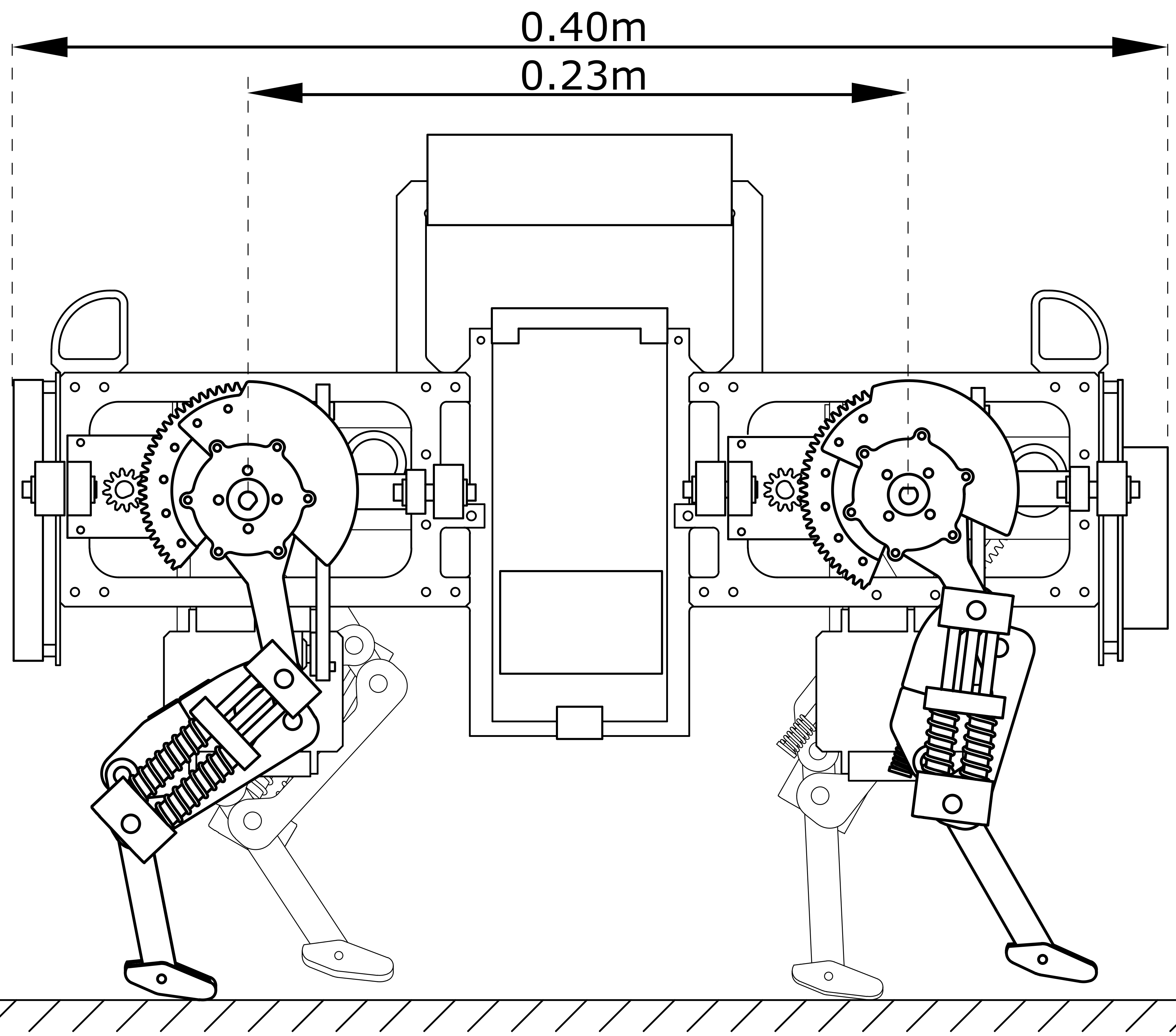}
\end{minipage}
\caption{
Oncilla robot: a) frontal, b) side view of a computer aided design,
with measures for the robot's
hip height during standing,
its overall height, width and length,
and its lateral and fore-aft distance between hip and shoulder joints.}
\label{fig:frontal-lateral}
\end{figure}

The robot was designed with mechanical improvements in mind compared to Cheetah-cub robot: increasing load carrying capabilities, incorporating an additional DOF per leg for efficient and fast turning, adding multiple modes of sensing, and adding batteries and on-board power for mobile application. Each additional feature increased the weight, from a \SI{1.1}{kg} Cheetah-cub robot to a \SI{5.1}{kg} Oncilla robot.
For Oncilla robot's larger weight, a stiffer leg spring was mounted, which in turn required a higher-torque actuator for the leg flexing mechanism. The LL actuator is mounted serially to the leg angle actuator, and shortens the leg through a cable (\cref{fig:load-scenario}). This design has positive consequences:
\begin{inparaenum}[\itshape a\upshape)]
\item The LL actuator is decoupled from external forces, such as sudden impacts. Hence, the mounted gearbox requires a lower safety rating, and can be designed with a lower module gear design.
\item During stance phase, parallel leg springs act passively, i.e. the design is a close approximation of the mechanism underlying the spring loaded inverted pendulum~\citep[SLIP template]{blickhan_spring-mass_1989}.
\end{inparaenum}
The sagittal leg angle actuator is directly attached to the proximal leg segment. The motor's long and narrow shape blocked mounting left and right leg motors in line. Instead we applied a non-symmetrical actuator placement, with legs relatively close to each other (\SI{0.14}{m}, \cref{fig:frontal-lateral} and Fig.\;S6).
The short distance leads to a short moment arm i.e. for diagonally touching legs, and is meant to reduce rolling motions.
Leg angle motor and hip axis are connected by a large module spur gear (\cref{fig:frontal-lateral}). 
Relative encoders are directly mounted to both brushless motor axes.
One absolute encoder is mounted at the main LA axis. Incremental motor encoders support precise motor control, and absolute encoders read leg segment positions, and leg spring loading.
\subsection{Hip adduction and abduction joint}
The added hip adduction and abduction (AA) actuation was installed for efficient turning.
The chosen AA actuator is a strong, position controlled RC servo motor (Kondo KRS 2350 ICS). It was selected for its compact design, high holding torque, and standardized control interface. The servo horn connects through a four bar mechanism with the leg, with a movement range of \ang{\pm 8} (\cref{fig:frontal-lateral}).
\subsection{Force sensors}%
Custom-made, two-axis force sensors were implemented proximally, within the AA suspension between trunk and legs (\circled{8}, \cref{fig:oncilla-photo-cad}). Force sensors were designed based on double cantilever bending beams, with foil strain sensing resistors in full bridge configuration. Sensors measure forces in vertical and fore-aft direction, in the trunk's coordinate system.
Oncilla robot's force sensors were not utilized for the gaits shown in this work. Instead, all gaits shown are generated in feed-forward mode. Force sensing is however necessary for either model-based control based on leg loading information, or non-model based controllers with i.e.\,reflex-like feedback~\citep{BIOROB-CONF-2008-003}. Because similar locomotion experiments are planned with Oncilla robot, custom designed force sensors are included here.

We considered two mounting places: distally, i.e. as feet, or proximally, between the robot's trunk and legs, as finally chosen.
When mounting force sensors distally, almost no effects from unsprung masses are measured, but direct contacts between leg and ground. Consequently, the resulting force signal requires less noise filtering. However, distally and foot-mounted sensors rotate with the foot frame. In case the direction of force is of interest for control, the sensor's orientation should be recalculated through the serial chain of trunk, leg segments, and foot sensor. Mechanically, a distal sensor placement moves the leg's center of mass further distally, especially for larger and more complex sensors. It also limits mounting spring-loaded feet. To cope with harsh touch down impacts, a miniaturized but also robust sensor design is required. 

In comparison, proximal force sensor mounts are relatively independent from leg and foot design. In our case sufficient mounting space was available proximally. Standard-sized, strain gage-based sensors were chosen and implemented. Drawbacks exists; proximally sensed force signals are influenced by mass and inertia of the moving leg, and data post-processing is required. Latter will introduce sensor signal delays. Such delays can present an obstacle especially for fast loop locomotion controllers. Proximal force sensors require no recalculation of the sensor's orientation, as they are fixed in the trunk frame.

\subsection{Electronics and PCB mounting}
Oncilla robot carries its actuators and sensors, a battery pack, and PCBs for motor control, power supply, and communication. Brushless motor driver boards were  custom designed, and each of four boards provides control and power for two brushless motors. One board weighs \SI{0.15}{kg}, and all are mounted low, at the robot's geometric center. This placement helps keeping the robot's COM below the virtual hip-shoulder axis---the robot's trunk 'hangs' in-between. The remaining PCBs are placed at the front of the robot (RB-110, main computing and communication), and at the rear end (power supply), for easy access. An inertia measurement unit (MicroStrain 3DM-GX3-35 IMU) is mounted above the motor driver boards. Further details on the electronic layout, components, operation system are available in the supplementary material (i.e. Fig.\,S9).

\section{Software}\label{sec:sw-arch}
In this section we describe software related concepts developed for the Oncilla robot; namely the communication bus connecting main control boards and periphery, the motor controller, Oncilla robot's software architecture, and the simulated Oncilla robot model in Webots.

\subsection{Communication layer and protocol}
\begin{figure}[h!]
\begin{center}
\includegraphics[width=0.7\textwidth]{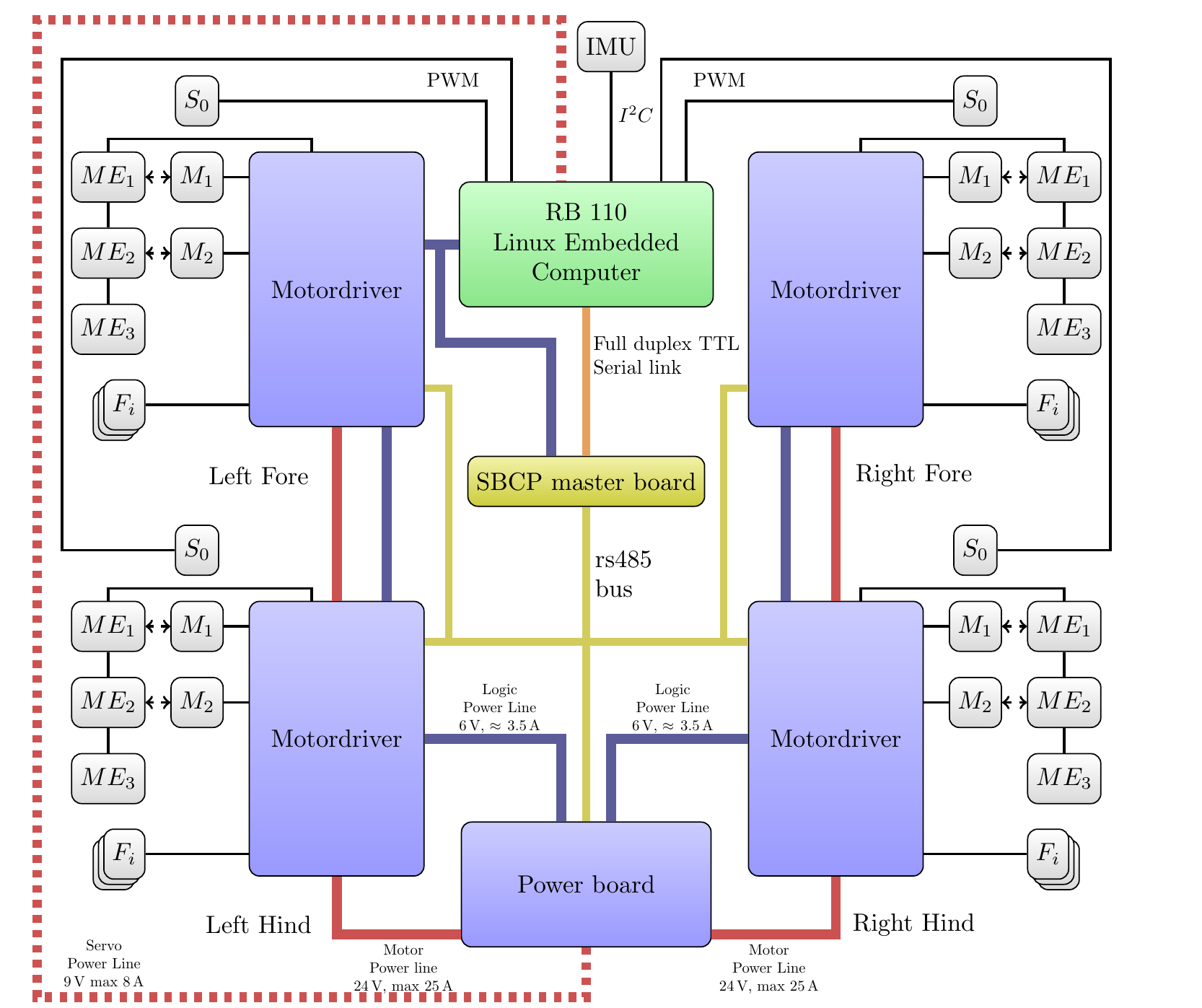}
\end{center}
\caption{
Schematics of Oncilla robot's electronics and communication network. 
Thick lines depict power supply for brushless motors (solid red), servo motors (red dotted), and logic (blue). 
Thin lines correspond to communication buses. 
All four legs feature: two brushless motors ($\mathrm{M}_1$, $\mathrm{M}_2$), one servo motor ($\mathrm{S}_0$), three absolute magnetic encoders ($\mathrm{ME}_1$, $\mathrm{ME}_2$, $\mathrm{ME}_3$), three strain sensor conversion channels ($\mathrm{F}_i$, two are used).}
\label{fig:electronics-communications}
\end{figure}

Communication between main electronic modules is performed over a RS-485 physical layer, organized in a master/slave point-to-multipoint half-duplex configuration (\cref{fig:electronics-communications}). The data link and network layer on the bus is implemented through a custom Simple Binary Communication Protocol (SBCP). Latter implementation is an extension of the Bioloid Dynamixel Communication Protocol version \num{1}~\citep{ax12}. The main modification is the change of the two byte packet preamble (value 0xFFFF). This preamble was separated; one single byte preamble (0XFF), and a ClassID byte to designate the device type we want to address, i.e. the motor driver board, power board, or master control board. By reserving the class value for Bioloid devices, Oncilla robot's hardware and driver are capable of directly incorporating these.

A major challenge in the SBCP design was the high communication bandwidth requirement. We initially aimed at a \SI{1}{kHz} control loop on the embedded computer level, this translates to a RS-485 baud rate of \SI{3.3}{Mbps}. Recently such high speed UARTs became available in embedded computers ($<\SI{12}{Mbps}$ at the moment), through an USB to serial Integrated Circuit (IC). However, USB bus communication is scheduled by \SI{1}{\ms} frame sizes, and a naive bus implementation would be limited to reach a device every \SI{2}{\ms}. Instead, we implemented a bus control data flow in a dedicated device (SBCP master board) controlled by a dsPIC33FJ128MC802 Digital Signal Processor (DSP), mounted after the USB to serial IC. This SBCP master board is able to:
\begin{inparaenum}[\itshape a\upshape)]
\item Handle a group of up to \num{8} combined packets sent over a full duplex UART connection.
\item Manage communication over the half duplex RS-485 bus with a latency of \SI{12}{\us} and maximal jitter of \SI{2.2}{\us}.
\item Detect slave timeouts and react accordingly with user customizable delays.
\end{inparaenum}
To reduce jitter and latency to a minimum, we utilized many optimization features of the dsPIC33FJ processor family such as Direct Memory Access (DMA), reduction of interruption stress on the processor, and anticipated packet precomputation and pre-buffering to reduce latency between packet responses. All implementation details are abstracted by a reusable library, SBCP-uc~\ref{tab:sources}. This new open source library enables fast development of slave device SBCP interfaces for this processor family.

\subsection{Motor control}
The custom designed motor driver boards implement a Proportional-Integral-Derivative (PID) controller for two brushless motors on a dsPIC33JF128MC804 DSP. It controls two A3930 motor driver ICs which are used to drive a three phase inverter made of six IRFR48Z MOSFETs. These ICs are able to limit the motor winding current.  
DSP PID controller parameters were hand tuned. In order to reduce motor jitter during position tracking, a velocity profile interpolation is implemented on the motor driver DSPs. The user sets the desired tracking frequency, and the tracking velocity is then computed on-board. Initially we observed trajectory tracking losses every several seconds. We found the cause in a combination of communication jitter and real-time clock frequency mismatch. Tested RB-110 boards had crystal frequency deviations of up to $+\SI{0.3}{\%}$, from their standard frequency. As a workaround, we implemented precise buffering and real-time clock re-synchronization heuristics on the motor driver DSP. The maximum tracking frequency is \SI{500}{Hz}, half of the DSP internal control loop (\SI{1}{kHz}).

\subsection{Software Architecture}
\begin{figure}[h!]
\begin{center}
\includegraphics[width=.98\textwidth]{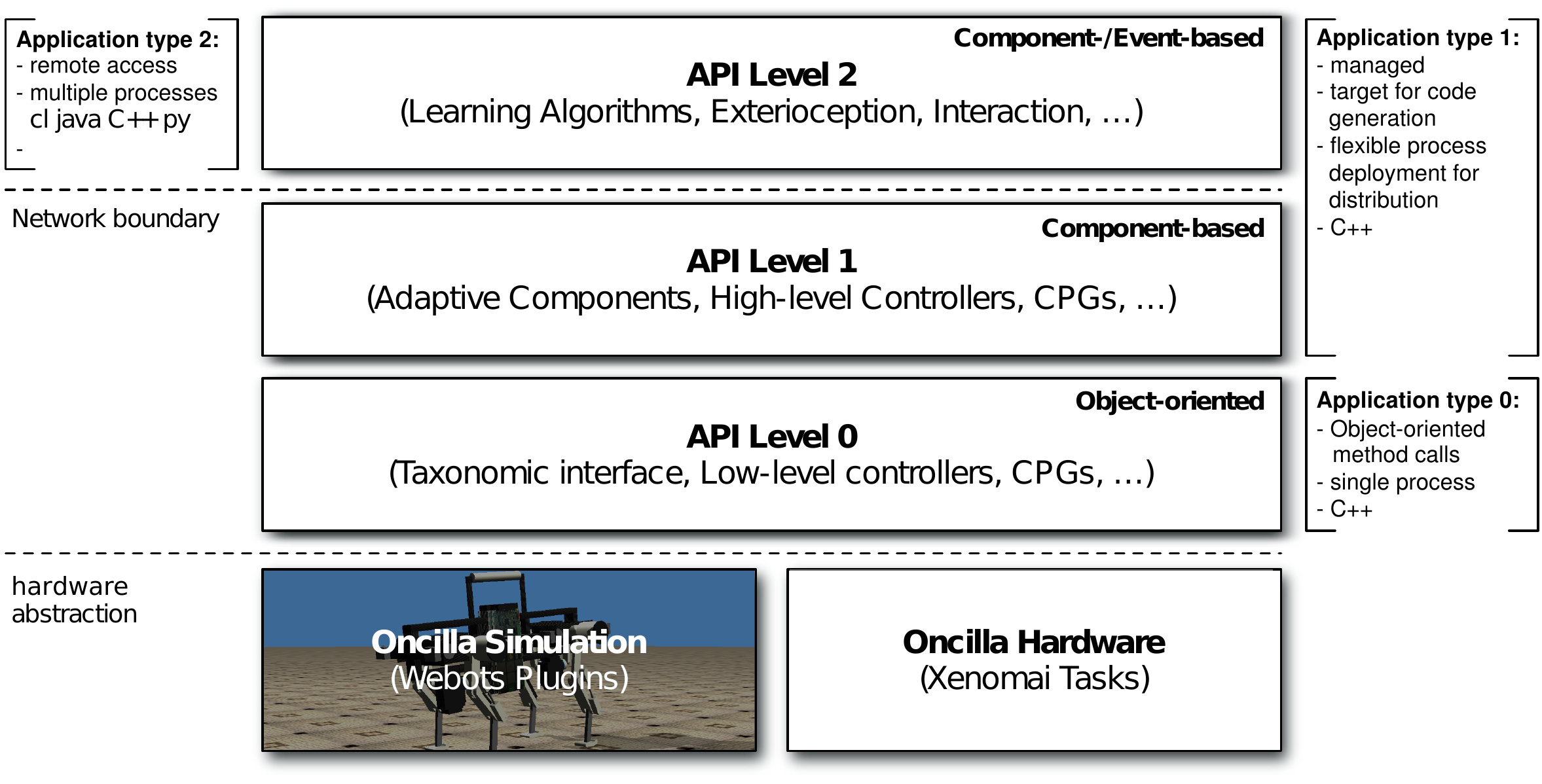}
\end{center}
\caption{
Schematic presentation of the three-layered Oncilla application programming interface (API). API levels \num{0} - \num{1} for local access, API level \num{2} for extended language and tool support over the network.}
\label{fig:sw:api-levels}
\end{figure}

The software architecture of Oncilla robot is based on two primary design decisions tied to the requirements of open research projects~\citep{nordmann_simtools_2012}: 
\begin{inparaenum}[\itshape a\upshape)]
\item It provides a common interface for the simulator and for the hardware. This allows for easy and fast transition of experiments between simulation and real-world experiments.
\item It provides a local interface for fast control loops running on the embedded PC, as well as a remote interface to allow more complex applications to control the robot over the network.
\end{inparaenum}
\Cref{fig:sw:api-levels} shows the software architecture that exposes the abstracted interface for both simulation and hardware at different application programming interface (API) levels according to application requirements.

\subsection{Simulation and Robot Interface}
One of the main design goals was to implement a common abstraction between hardware and simulation, with binary compatibility, to facilitate an easy transfer between hardware and simulation. The abstraction also allows to exchange the currently Webots-based~\citep{webots} simulation back end. 
Here we chose Webots software as our multi-body simulation software for Oncilla robot.
Since the lowest API level \num{0} is implemented for the simulation and the hardware back end, applications implemented against any of the API level \num{0}-\num{2} can be switched transparently between simulation and hardware (\cref{fig:sw:api-levels}). By providing a common abstracted interface for both simulation and hardware, it enables fast and easy transfer of experiments between these two domains. It is also possible to replay real experiment recordings (e.g.~joint angles) in simulation, and vice-versa.

\subsection{Local and Remote Interface} 
The Oncilla interface is accessible through a local C++ interface (API levels \num{0} and \num{1}, \cref{fig:sw:api-levels}) and remotely via the open-source middleware for extended language and tool support (API level 2). The low-level sensors and actuators are locally accessible through a C++ interface, using multiple inheritance to expose the node taxonomy. The local interface enables light-weight applications with fast sensor feedback running on the embedded PC without dealing with network latency.
The Oncilla interface is also remotely available by the open-source middle-ware Robotics Service Bus (RSB) with C++, Java, Python, and Common Lisp bindings and therefore allows applications in all four languages~\citep{Wienke2011}. This enables extended tool support, e.g. external logging, monitoring and recording of experiments. It supports more complex and computationally expensive applications to run on distributed PCs and control of the robot over the network. This was successfully applied and demonstrated in experiments that were specified in domain-specific languages. The experimental source code was automatically generated, as well as machine learning applications that can not run on the embedded PC due to resource limitations~\citep{nordmann_simtools_2012}.

\subsection{Webots Model of Oncilla robot}
A simulated model of Oncilla robot was created in Webots~\citep{webots}. It is based on the mechanical properties extracted from Oncilla robot's computer aided design (CAD) model, i.e.\;weight, center of mass, inertia, link dimension, and spring constant.
The motivation to use a physics engine intended for games such as Open Dynamic Engine (ODE, physics engine of Webots) is a balance between simulation speed and required simulation accuracy. 
Internal kinematic loops in Oncilla robot's pantograph leg, and its asymmetric actuation can be expressed as a constraint in the underlying Linear Complementary Problem (LCP).  There is a drawback to this approach; Open Dynamic Engine favors stability over accuracy, which results in a poor constraint resolution. Likely, such a simulator is well suited for fast prototyping, but less suited for on-board, model-based control.  We used the default Webots actuator model for the proximal joints (hip and shoulder). This model applies a PID controller, with a maximum output velocity and torque limited by the theoretical motor limits, i.e. a maximum torque of $\SI{53.5}{mNm} \cdot \num{84} = \SI{4.5}{Nm}$ and a maximum speed of $\SI{16300}{rmp} / \SI{60}{s} \cdot 2 \cdot pi / 84 = \SI{20.3}{rad/s}$ (\SI{53.5}{mNm} constant motor torque, overall gear ratio $n=\num{84}$, \SI{16300}{rpm} no-load motor speed).  For the leg's flexion and extension DOF we are directly manipulating the LCP constraint to control the mechanical stop of the diagonal spring.  Since there is no easy way to specify the maximum torque or linear force in ODE to satisfy a given constraint, we transformed this force constraint into a velocity constraint using a linear motor model with internal resistance.  We simplified and assumed that the motor torque required for actuating the leg length cable was caused by the compressive force of the diagonal spring. In reality, this torque is an upper bound since external forces could induce additional leg flexion.  We further limited the maximum speed at which motors move to the mechanical stop constraint for the next simulation step, proportionally to the instantaneous torque. This limit is ranging from maximum speed with no torque required, to zero speed at maximum motor torque requested. The resulting model corresponds to a simple, linear motor model. Finally an implementation of the C++ interface level \num{0} back end was developed. It provides the user with the ability to use the same API to seamlessly switch control between either the real hardware or the simulated robot.

\section{Experiments and experimental results}
\label{sec:experiments}
This section describes experimental results with the hardware Oncilla robot utilizing the here described open-loop controller for straight level locomotion indoors, locomotion descending and ascending slopes, and turning strategies with and without AA joints. Links to videos of Oncilla robot locomotion are provided in Table S1 (supplementary material). Further, we provide experimental results with the simulated Oncilla robot running in Webots. 
Advanced experiments with Oncilla robot's closed loop control framework are documented in \cite{mos-thesis}.
\subsection{CPG controller for straight locomotion}\label{label-locomotion-controller}
For the robot's locomotion control, we applied morphed oscillators~\citep{ajallooeian2013general}, to implement a Central Pattern Generator model~\citep{ijspeert2008central}. Morphed oscillators are nonlinear oscillators which can encode arbitrary limit cycles defined as phase-dependent functions. Given a desired joint trajectory, a morphed oscillator can be implemented to encode this trajectory as a stable limit cycle. This provides a smooth trajectory generator with the capability of feedback integration. A morphed oscillator utilizes a simple oscillator as base and morphs it to obtain the desired limit cycle behavior. Here we use a unit radius amplitude controlled oscillators as base:
\begin{eqnarray}
  \dot{\theta}_{i} &=& \varOmega_i \label{eq:theta} \\
  \dot{r}_{i} &=& \varOmega_i f_i'(\theta_i) + \gamma \left( f_i(\theta_i) - r_i \right) + \xi_i \label{eq:r} \\
  \varOmega_i &=& \omega + \sum_{j = 1}^{N} c_{ij} \sin(\theta_j - \theta_i - \phi_{ij}) \label{eq:coupling}
\end{eqnarray}
where $\theta_i$, $\varOmega_i$ and $r_i$ are the phase, the coupling dynamics, and the radial output of the $i$th oscillator, respectively.  $\gamma$ is the rate at which the dynamics converge to the limit cycle, $\omega$ is the locomotion frequency multiplied by $2\pi$, and $c_{ij}$ and $\phi_{ij}$ are the coupling strength and phase difference between the $i$th and $j$th oscillator. The phase difference is exploited to implement inter-joint coordination. $f_i(\theta)$ defines the shape of the limit cycle of the $i$th oscillator and $f_i'(\theta) = \partial f_i(\theta) / \partial \theta$. $\xi_i$ is the additive feedback. It can be designed through strategies explained in~\cite{ajallooeian2013central, ajallooeian2013general}. $r_i$, the time-integration of $\dot{r}_{i}$, is the joint angle reference for the $i$th DOF.

To design locomotion gaits, we define foot trajectories with respect to the hip frame, similar to~\cite{maufroy2010integration}. We use simplified closed-form inverse kinematics to convert those to joint trajectories. These joint trajectories define $f_i(.)$ functions. The definition of $\phi_{ij}$ is gait dependent, for example for a trot gait $\phi_{ij} = \pi$ for adjacent hips, and $\phi_{ij} = 0$ for diagonal hips. Finally, $c_{ij}=5$ for all the oscillators.
We applied solely open-loop gaits in this work. However, Oncilla robot's hardware and software architecture allows to utilize the robot's internal sensors and apply closed-loop control, i.e. with reflexes and posture control~\citep{mos-thesis}.

\subsection{Level trotting}
\begin{figure}[h!]
\begin{center}
\includegraphics[width=.98\textwidth]{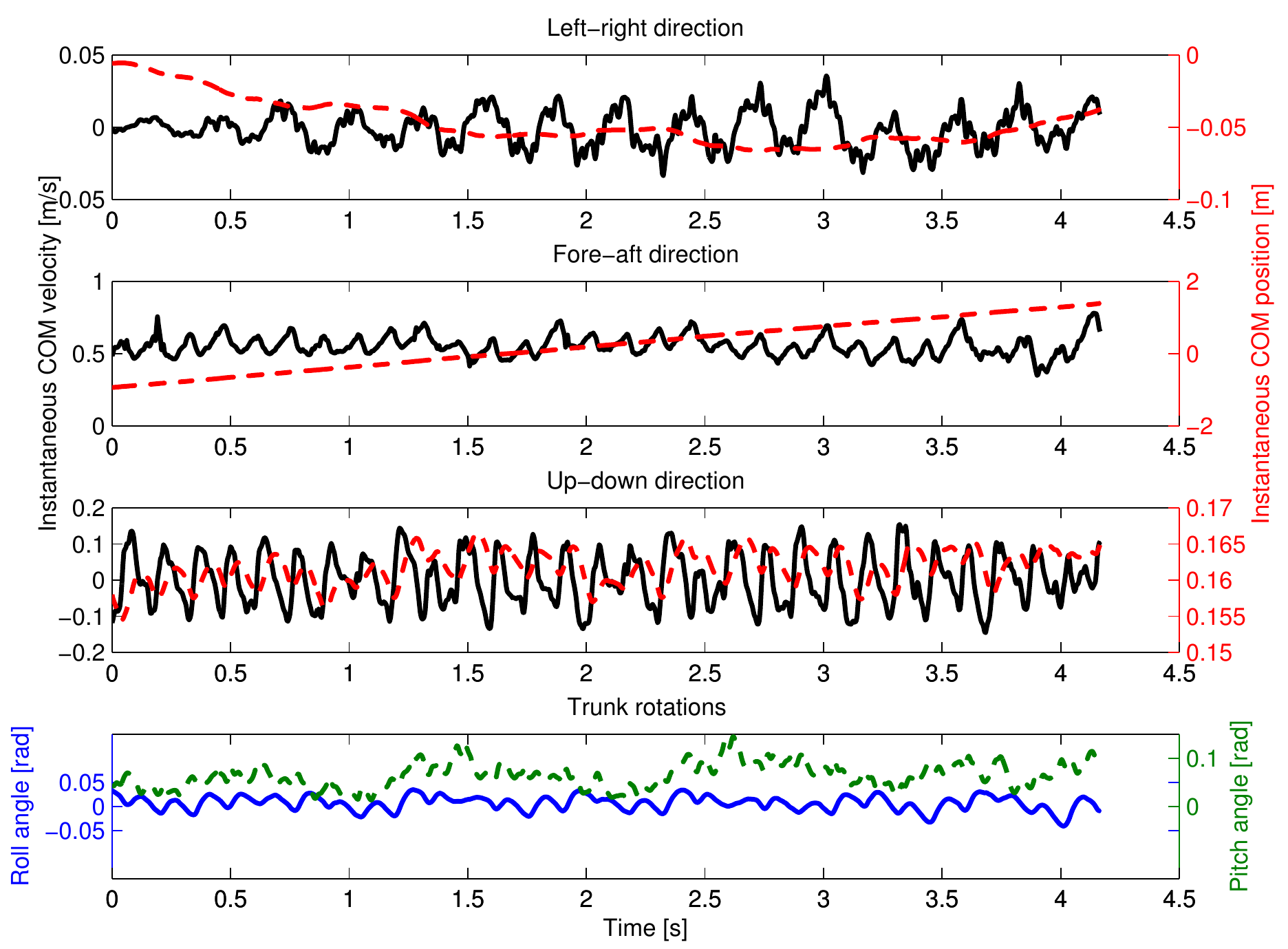}
\end{center}
\caption{
Hardware experiment. 
Oncilla robot's center of mass (COM) position and velocity are plotted over time, at locomotion cycle frequency of \SI{3.5}{Hz}. 
Shown are instantaneous COM position (red dashed) and velocity (black full). 
Components are sorted by their left-right, fore-aft, and up-down components. 
In this example, the robot trotted with an average speed of \SI{0.55}{\mps} and a maximum instantaneous (peak) forward speed of \SI{0.78}{\mps}. 
Vertical displacement of the robot's COM was \SI{\pm 5}{mm}, average hip height \SI{0.16}{m}. 
Average roll angle around fore-aft axis was \SI{\pm 0.02}{rad}. 
The average pitch angle around left-right axis was \num{0.06} \SI{\pm 0.04}{rad}.
}
 \label{fig:COM-speed}
\end{figure}

\Cref{fig:COM-speed} shows experimental data of Oncilla robot trotting forward at an average speed of \SI{0.55}{m/s} ($\mathrm{Fr}= \num{0.19}$), on level terrain, on a standard laboratory surface and with the CPG controller described in the previous section, in open-loop mode. Kinematic data was recorded at \SI{240}{Hz} with a motion capture system~\citep{naturalpoint}. Position and velocity data of the robot are shown, ordered by its left-right, fore-aft, and its up-down direction. The robot's center of mass (COM) oscillated vertically about \SI{\pm 5}{mm}, left-right velocity stayed in the range of \SI{\pm 0.05}{m/s}. Peak forward speed of this recording was \SI{0.78}{m/s}. The recorded roll angle during trotting stayed symmetrically around $0 \pm \SI{0.02}{rad}$ ($\SI{0}{\degree} \SI{\pm 1.2}{\degree}$), the pitch angle of the robot oscillated around $0.06 \pm \SI{0.04}{rad}$ ($\SI{3.4}{\degree}\pm\SI{2.3}{\degree}$).

In further tests, Oncilla robot's best forward trotting speed at \SI{3.5}{Hz} was $\mathrm{v}=\SI{0.63}{m/s}$. This is equivalent to \num{2.7} body lengths per second, with a body length of \SI{0.23}{m} (shoulder-to-hip distance, \ref{fig:frontal-lateral}).
The best average speed backwards was \SI{0.78}{m/s} (\SI{3.4}{BL/sec}), at \SI{4}{Hz} locomotion frequency. 
Backwards trotting with Oncilla robot showed less slippage, leading to a larger effective stride length.
Forward locomotion frequencies above \SI{3.5}{Hz} yielded no speed gain, due to slippage. 
Links to videos of the robot trotting both indoors and outdoors (gravel, step-down, flat terrain) are provided in Table S1.

Cost of transport (COT) was measured on the tethered robot ($\mathrm{m}=\SI{4.5}{kg}$). Stand-by power consumption with no actuator movement was subtracted from all runs (\SI{19.6}{W}), the remaining power consumption (P) was used to calculate the cost of transport~\citep{tucker_energetic_1970}:
$\mathrm{COT} = \frac{\mathrm{P}}{\mathrm{mgv}} = [\si{J/(Nm)}]$
with $\mathrm{g}=\SI{9.81}{\meter\per\second^2}$, and average speed $\mathrm{v}$ over at least \num{4} cycles. Although COT is often given without units, we use [\si{J/(Nm)}], to avoid mix-up with COT values given in [\si{J/(kg m)}]. Froude numbers are calculated as $\mathrm{Fr}=\mathrm{v}^2 / (\mathrm{gl})$, with $\mathrm{l}=\SI{0.16}{m}$ standing hip height. Duty factor values are given as ratio of leg stance to cycle time.

\begin{figure}[h!]
\begin{center}
\includegraphics[width=10cm]{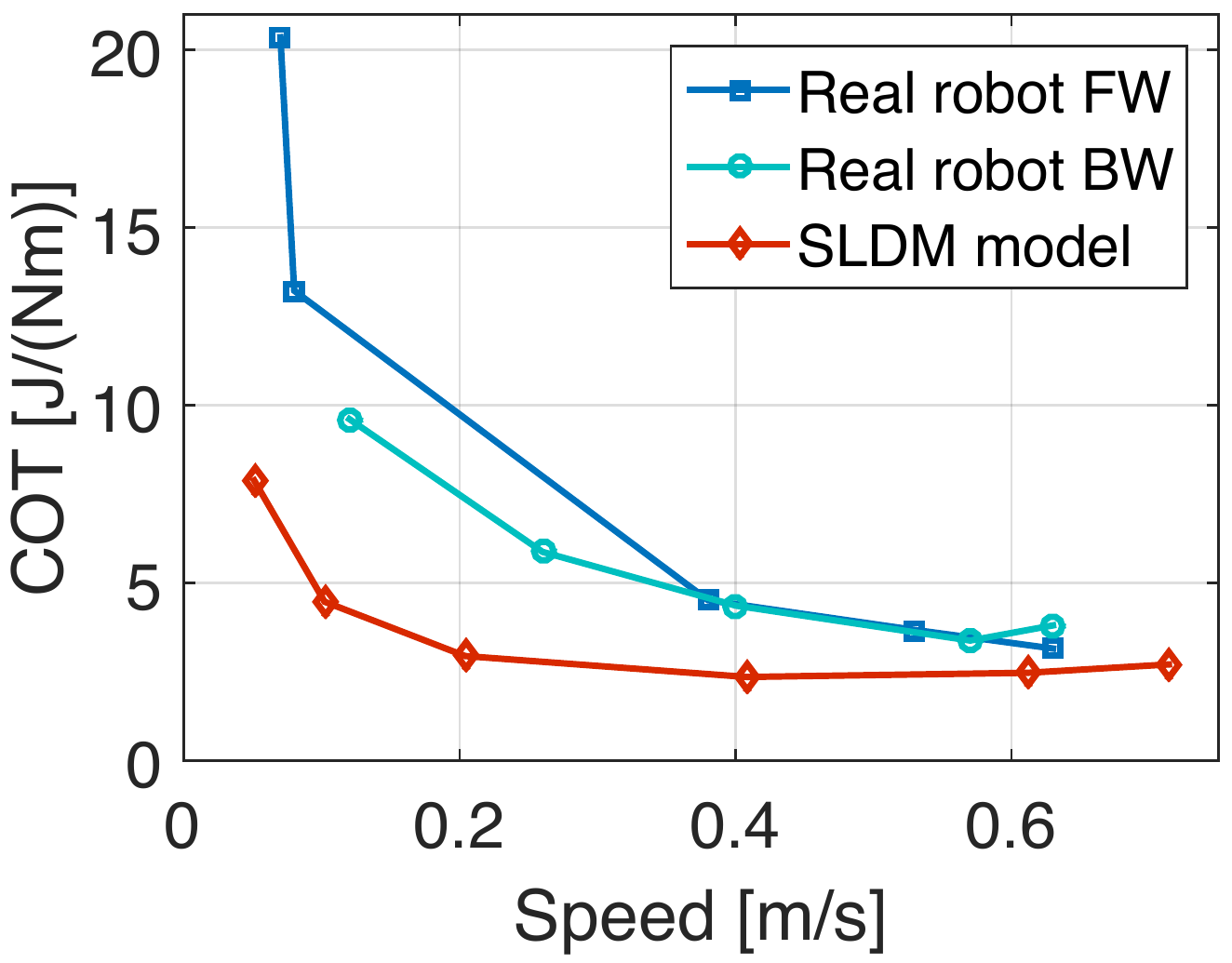}
\end{center}
\caption{
Hardware experiment and results from simplified-load dynamical motor (SLDM) model. 
Plots show the cost of transport (COT) of the real Oncilla robot and the SLDM model over different speeds. 
The full power consumption minus the stand-by robot power consumption (\SI{19.6}{W}) was used for the COT calculation. 
Red diamond (SLDM model) marks show the estimated COT values calculated prior to the construction of Oncilla robot based on a simplified, dynamically articulated robot model. 
Dark blue data points show the COT-speed values for the real Oncilla robot during level trotting in forward (FW) direction. 
Round marks indicate the hardware robot's COT during backwards (BW) level trotting. 
The FW locomotion shows a higher COT up to a speed of \SI{0.4}{\mps}, compared to the BW locomotion. 
The SLDM model continuously underestimates the real robot's COT, but provides a good estimate of the asymptotic decline of COT over speed. 
The best recorded COT with the real Oncilla robot is \SI{3.2}{J/(Nm)} at \SI{0.63}{\mps}, during FW trotting.}
\label{fig:COT-3cases}
\end{figure}

\Cref{fig:COT-3cases} displays recordings for speed and cost of transport (COT) for the simplified modeled ('SLDM',~\cref{sec:load-scenario}) and the hardware (`real') Oncilla robot, during forward (square markers) and backward (round markers) trotting locomotion, over \num{5} tested speeds. At very low speed (\SI{0.07}{m/s}) the robot trotted forward with a high COT of \SI{20.4}{J/(Nm)}. It reached its best forward COT at its maximum recorded speed on level terrain: \SI{0.63}{m/s} with a COT of \SI{3.2}{J/(Nm)}. Backwards locomotion was more efficient at low speed, with a COT of \SI{9.6}{J/(Nm)} at $\mathrm{Fr}=\num{0.01}$. At higher backwards speed (\SI{0.63}{m/s}) the robot trotted with a COT of \SI{3.8}{J/(Nm)}. 

\cref{fig:COT-3cases} shows the SLDM model underestimates the real robot's COT. Considering its simplicity, the SLDM model does approximate the hardware robot's COT well, with \SI{2.8}{J/(Nm)} at \SI{0.71}{m/s}. The model also qualitatively captures the asymptotic decrease of COT over speed. 

\subsection{Slope up and down locomotion}
\begin{figure}[h!]
\begin{minipage}[b]{.5\linewidth}
\centering\includegraphics[width=6cm]{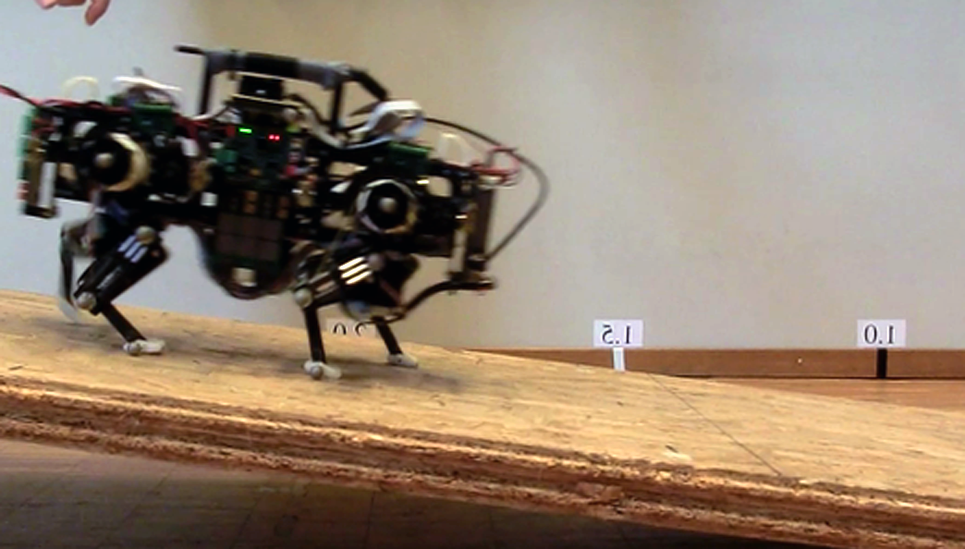}
\end{minipage}
\begin{minipage}[b]{.5\linewidth}
\centering\includegraphics[width=6cm]{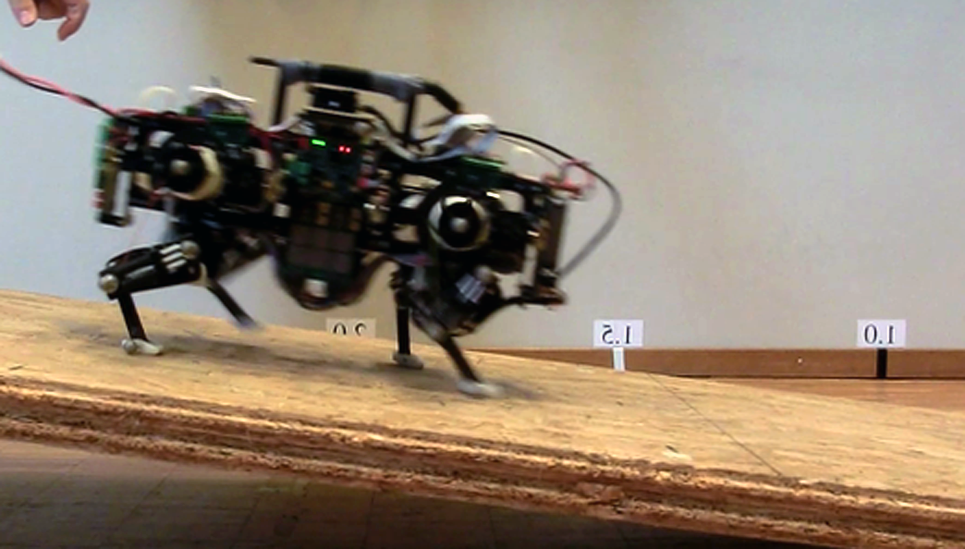}
\end{minipage}
\begin{minipage}[b]{.5\linewidth}
\centering\includegraphics[width=6cm]{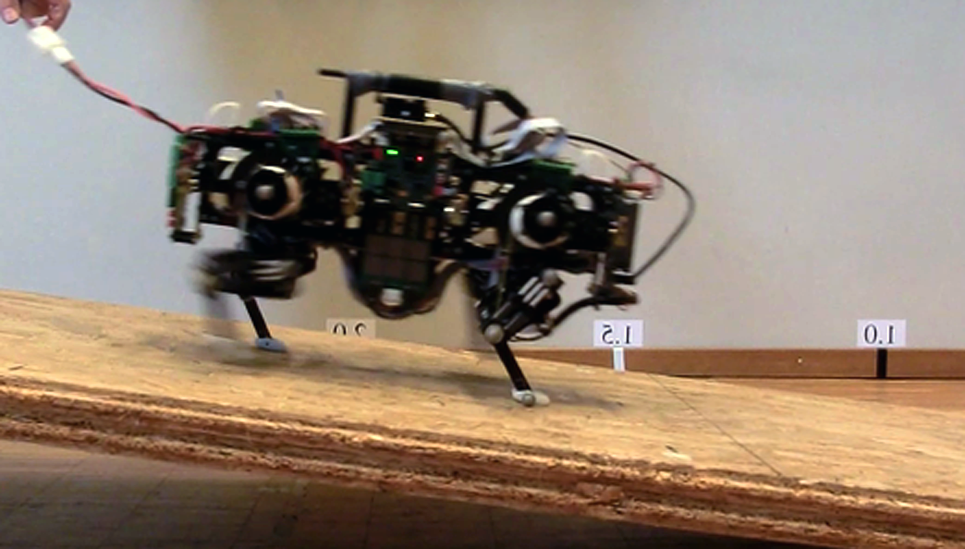}
\end{minipage}
\begin{minipage}[b]{.5\linewidth}
\centering\includegraphics[width=6cm]{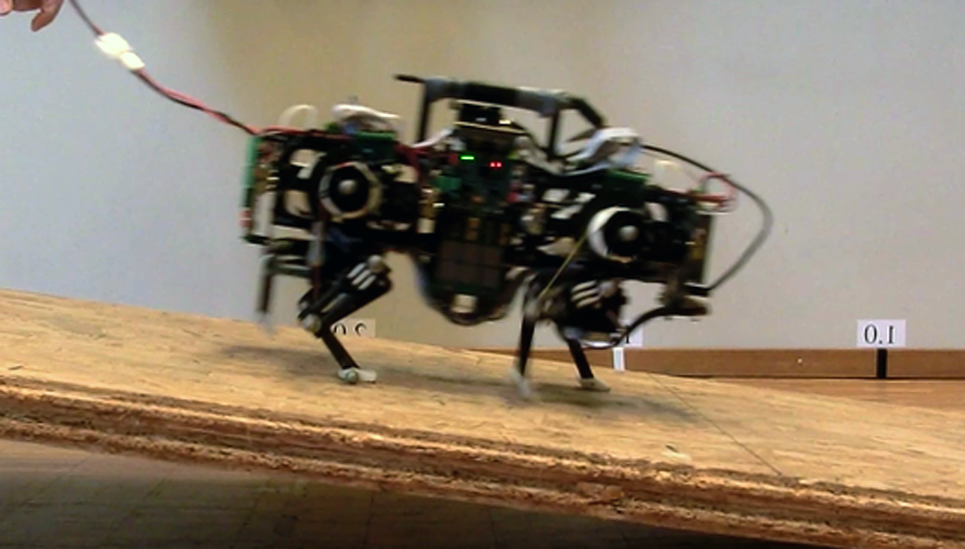}
\end{minipage}
\caption{
Hardware experiment: snapshots of Oncilla robot descending a slope in forward direction. Further tests were performed with the robot going up the slope, and by letting the robot climbing and descending during backwards locomotion. At steeper slopes, Oncilla robot showed excessive slippage when climbing the slope head on (\cref{tab:inclined}). Generally, the robot performed better when locomoting backwards. Snapshots here are flipped horizontally, for reading convenience.}
\label{fig:snapshots-slope}
\end{figure}

\Cref{tab:inclined} illustrates results for locomotion on inclined surfaces, and \cref{fig:snapshots-slope} depicts a slope descending locomotion in  forward direction. All slope experiments were conducted without changes to the standard level-trotting locomotion controller, or changes to the hardware. 
This allows better comparison between level and slope trotting, although adapting the foot friction and gait patterns, and applying closed loop control would improve the robot's speed~\citep{ajallooeian2013central}. For the experiment documented in \cref{tab:inclined}, Oncilla robot trotted forward and backward, level up and down slopes up to \ang{10}.

For inclinations of more than \ang{4}, the robot could only climb slopes when going backwards. 
The highest inclination climbing was recorded at \SI{10}{\degree}, with the robot going backwards at \SI{0.25}{m/s}, at \SI{0.4}{m/s} commanded speed. 
The robot trotted forward onto slopes of \SI{4}{\degree}, with a speed of \SI{0.15}{m/s}. 
Down, the robot kept the commanded speed of \SI{0.4}{m/s} when pointed forward. 
Its speed increased by \SI{5}{\%} when going down backwards (\SI{0.42}{m/s}).
From video footage we observed that runs with strong speed deviations coincided with strong robot feet slippage. 

\subsection{Turning maneuvers} 
We utilized two strategies to implement turning. For the first method (adductor/abductor amplification: 'AA-amp'), a sine-wave was embedded into the oscillator nodes controlling the adduction and abduction joints. Turning was achieved by setting the amplitude of the fore and hind AA joints ($a_{l, AA}$) with opposite signs, proportional to the desired turning rate ($\Delta \psi_{\mathit{yaw},\mathit{des}}$). 
In AA-amp, turning time is only a variable of AA amplitude. 
\begin{eqnarray}
f_{l, \mathit{AA}} &=& a_{l, \mathit{AA}} \sin(\theta_{l, \mathit{AA}}) ~~~ l = 1..4 \label{eq:aaf}\\
a_{l, \mathit{AA}} &=& \lambda \Delta \psi_{\mathit{yaw}, \mathit{des}} \label{eq:aaamp} \\
\lambda &=& \begin{cases}
+1  & l \in \{\mathit{LF}, \mathit{RF}\} \\
-1  & l \in \{\mathit{LH}, \mathit{RH}\} 
\end{cases}
\label{eqn:turning-1}
\end{eqnarray}
 
For the second method (asymmetrically shortening stride length: 'ASL turning'), step length between the left and right leg was modified in order to implement turning, without the use of AA joints. We implemented a turning strategy typically used for two-wheeled mobile robots; the step length was shortened for the legs on one side of the body, to turn in the same direction. The robot turns in-place if step lengths are equal, but with opposing signs. 
\begin{equation}
a_{l,\mathit{asym}} = \begin{cases}
2 \varpi + 1  & \varpi < 0 ~\&~ l \in \{\mathit{LF}, \mathit{LH}\} \\
1  & \varpi > 0 ~\&~ l \in \{\mathit{LF}, \mathit{LH}\} \\
1  & \varpi < 0 ~\&~ l \in \{\mathit{RF}, \mathit{RH}\} \\
1 - 2 \varpi  & \varpi > 0 ~\&~ l \in \{\mathit{RF}, \mathit{RH}\} 
\end{cases}
\label{eqn:turning-2}
\end{equation}
 $a_{l,\mathit{asym}}$ is the step length amplifier, for leg $l$, and $\varpi$ is the turning factor.
This approach produced a small amount of slippage because the ground-contacting legs (diagonal pair of legs, during trot gait) are not on the same axis. Note that in this approach the turning time is a variable of both the commanded forward velocity and the turning factor.

AA amplification was used for most of the turning cases, however fast turning with this method resulted in considerably larger forces at AA joints. The ASL method was mostly used to turn when the ground was uneven. This method does not depend on lateral foot movement, and therefore poses no danger of sideways stumbling against obstacles. Quantitative results from experiments with these two turning strategies are provided in \cref{tab:turning}. They show that AA-amp allows to turn on the spot, taking \SI{10}{s} for a full turn. The AA strategy leads to small speed losses during turning, between \SI{20}{\%} to \SI{30}{\%}. AA-amp turning radius was between \SI{0.23}{m} and \SI{0.46}{m} and depended on turning speed. 
In comparison, the ASL method showed higher speed losses at turning. The best parameter configuration led to a \SI{52}{\%} speed loss, at a turning radius of \SI{0.5}{m}. The smallest ASL turning radius recorded was \SI{0.03}{m}. Turning maneuvers, including on-spot turning, are available as supplementary videos (Table S1).

\subsection{Leg and force sensor characterization}
\begin{figure}[h!]
\begin{minipage}[b]{.5\linewidth}
\centering\includegraphics[width=.98\textwidth]{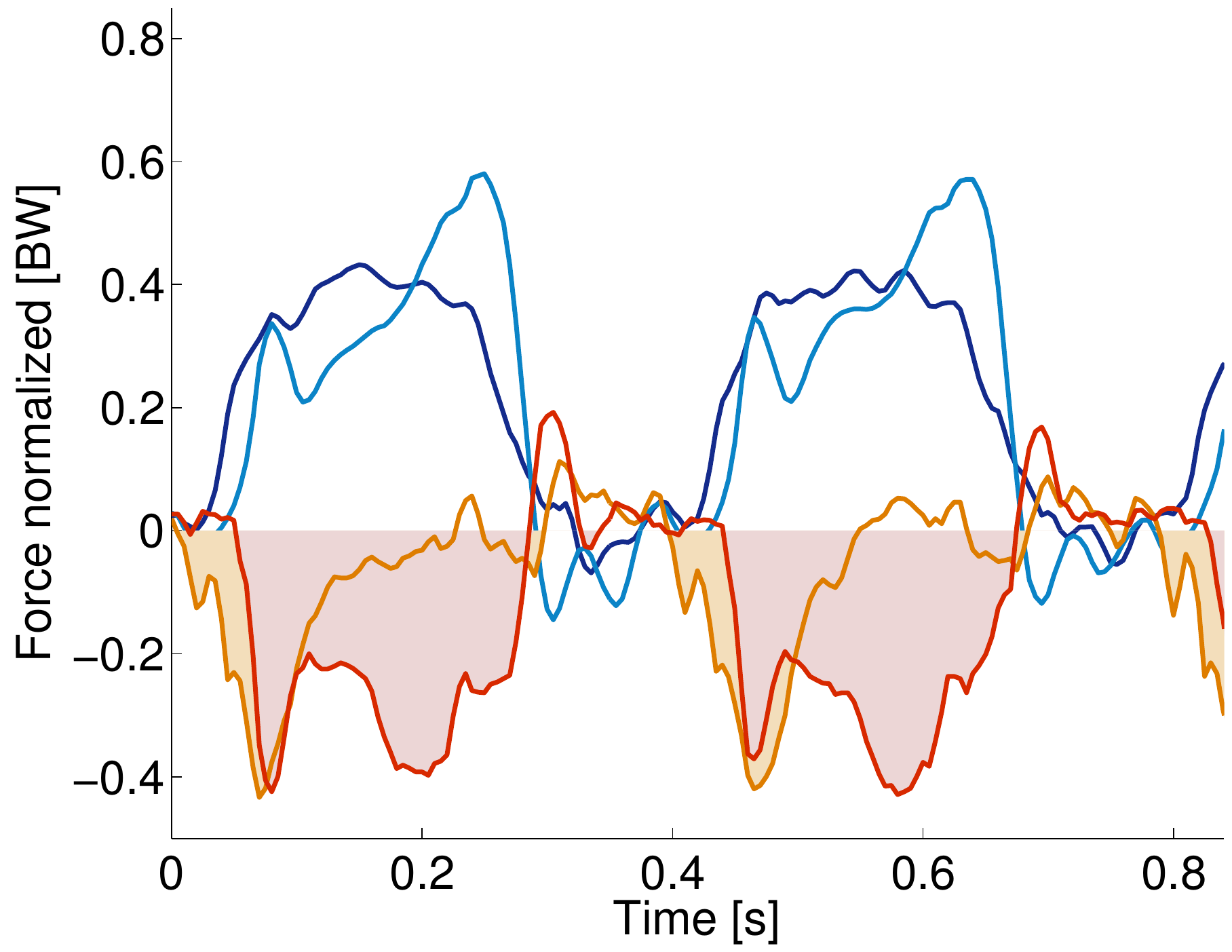}
\end{minipage}
\begin{minipage}[b]{.5\linewidth}
\centering\includegraphics[width=.98\textwidth]{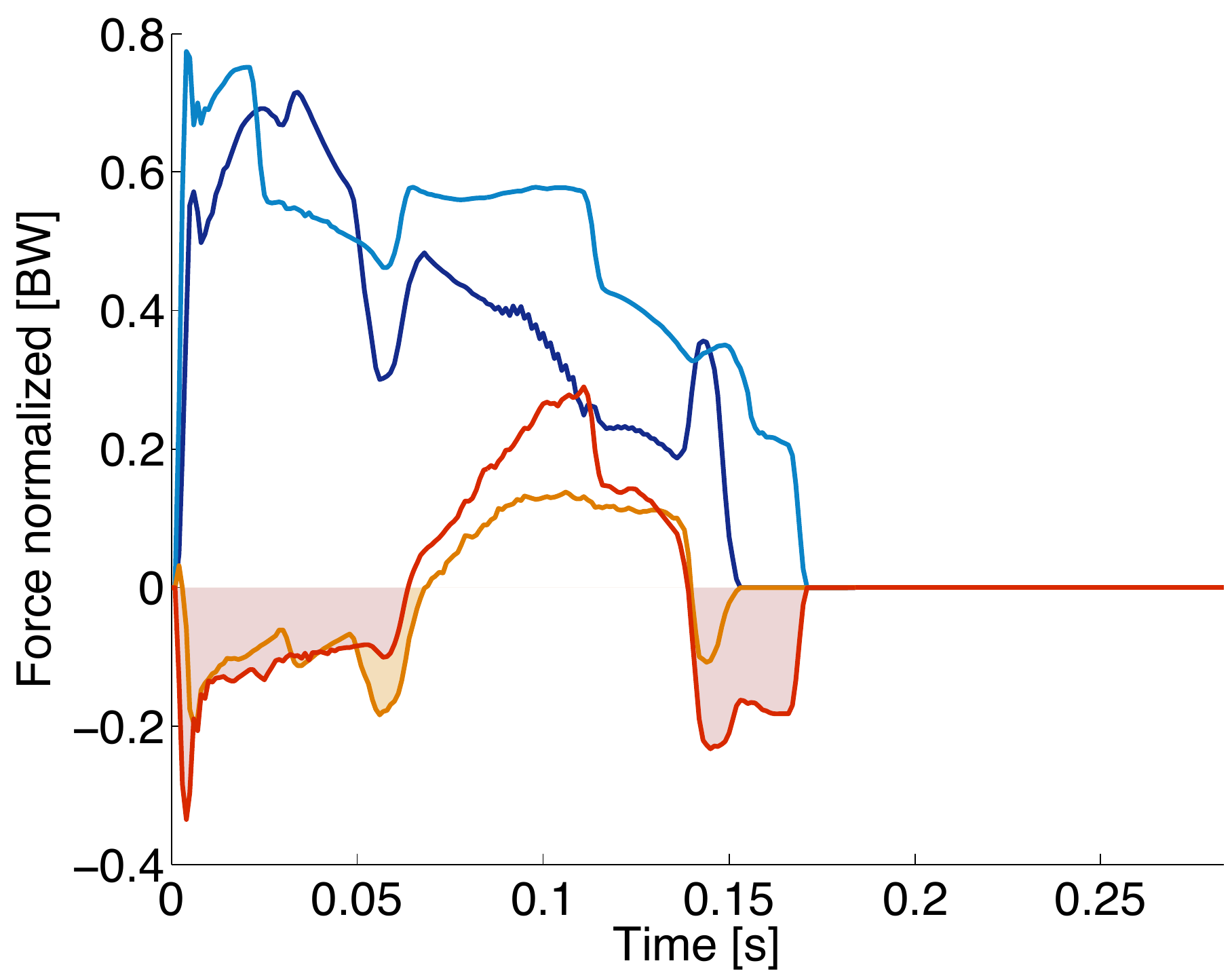}
\end{minipage}
\caption{
Oncilla robot's force sensor signals for 
a) the hardware robot, and 
b) its simulated Webots model. 
Forces are given in multiples of body weight (BW). 
Recorded are hind right (HR) and left front (LF) legs. 
Vertical forces $F_\mathrm{ver,LF}$: dark blue line, $F_\mathrm{ver,HR}$: light blue line. 
Horizontal forces $F_\mathrm{hor,LF}$: orange line, $F_\mathrm{hor,HR}$: red line. 
The hardware robot gait is a \SI{2.5}{Hz} trot, the simulated robot trotted at \SI{3.5}{Hz}. 
Hardware data was post-processed by an offset correction, and a \SI{18}{Hz} low pass filter. 
The hardware experiment indicates vertical forces around \SI{0.5}{BW}. 
Time-wise integration of horizontal forces over stance phase would be zero at constant velocity trotting. 
Hardware recorded force data however shows in-sum negative impulse. I.e.\,the data indicates a decelerating robot, while the actual robot was trotting at quasi continuous speed. We assume that offsets are created by internally deflected mounting brackets, around the horizontal force sensors. Positive and negative components of horizontal forces extracted from Webots are about equal.}
\label{fig:force-sensors}
\end{figure}

The recorded sensor data of the robot's proximal, trunk-mounted force sensors shows mixed results~(\cref{fig:force-sensors}). Especially the horizontal force signal is sensitive to parasitic stresses and strains of the mounting brackets, caused by forces between legs and trunk. To keep its weight low, the robot is designed from lower stiffness materials such as 3D printed plastic. As a consequence, the AA suspension bracket deflects under load, and influences forces sensor readings. In the horizontal force signal, this appears as high-frequency noise.
The robot's vertical force sensor showed good signal quality. After calibration and post-processing, vertical reaction forces can be extracted (\cref{fig:force-sensors}), which can be utilized for future feedback-based controllers.

\begin{figure}[h!]
\begin{center}
\includegraphics[width=10cm]{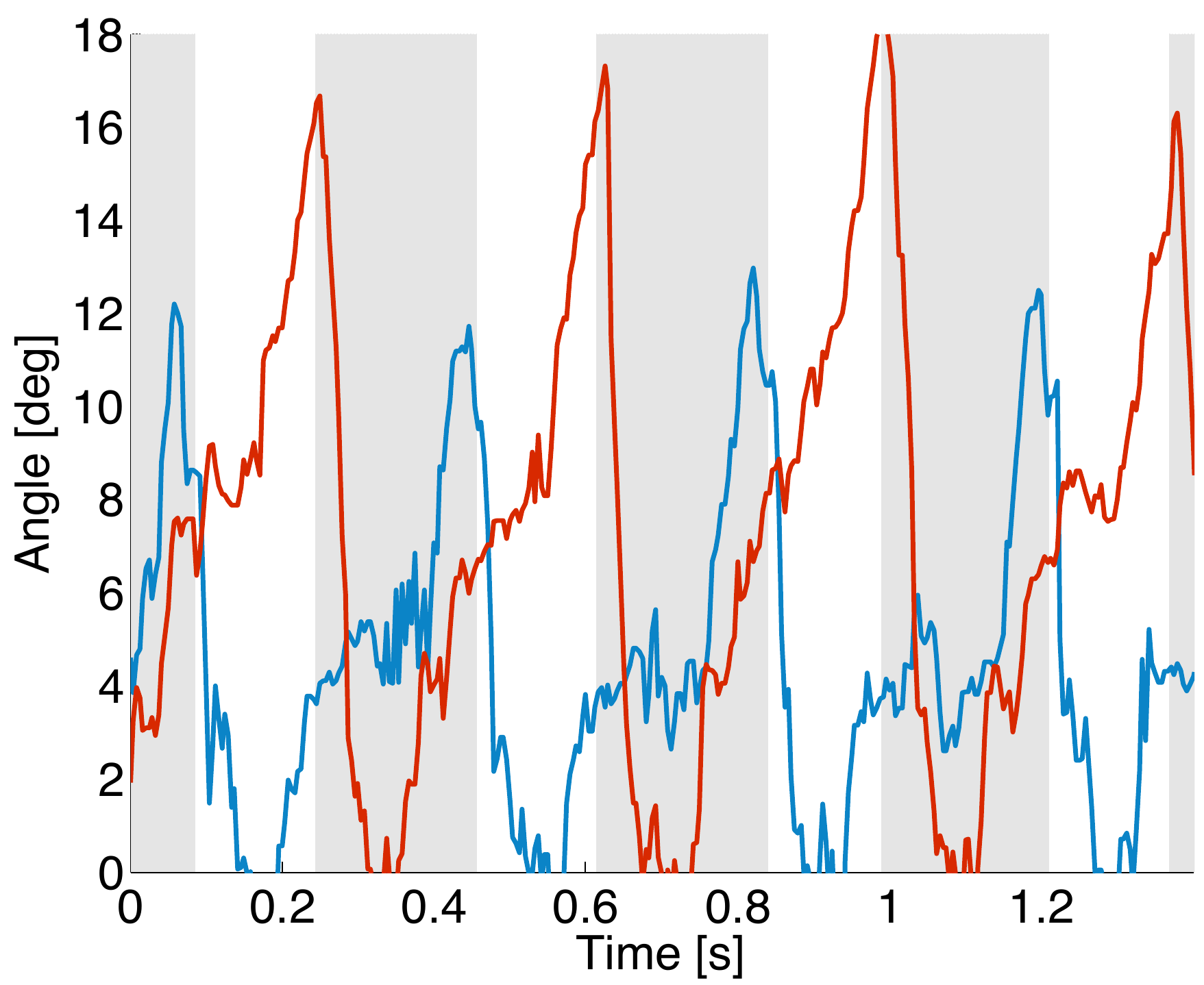}
\end{center}
\caption{
Hardware experiment. The difference between knee and ankle joint position ($p_\mathrm{knee}-p_\mathrm{ankle}$) is plotted, for the front left (blue line) and a hind left (red line) leg. 
Stance phases of the front leg are shown with a gray background, white background indicates the swing phase. 
Leg loading information is imperfect, and joint friction leads to delayed and damped joint movements. 
This overestimates loading contact times, but can be filtered. 
The here shown data is not filtered.}
\label{fig:sensor-knee-ankle}
\end{figure}

Stance phase timing, and to some extend leg loading can be estimated by observing the difference of leg joint deflection between knee joints, and spring loaded ankle joints ($p_\mathrm{knee} - p_\mathrm{ankle}$). \cref{fig:sensor-knee-ankle} shows the resulting angular signal, with joints being charged periodically at each stance phase. If required, this angular difference signal can be multiplied by the ankle joint stiffness, creating a source for ankle joint torque sensory feedback (not shown here). Importantly, this method is comparatively cheap, both from a hardware and computational perspective. It requires no additional hardware sensing framework, while providing joint angular, stance timing, and loading information.

\subsection{Oncilla Webots Simulation}
The Webots~\citep{webots} simulated Oncilla robot allows one to test control algorithms and locomotion scenarios without access to the real hardware. The controller can then be transferred to the robot, with the help of Oncilla robot's software architecture. Certain limitations for a direct transfer of control parameters exist ('one-to-one transfer') because of the 'reality gap' between simulated and real robot, those are also explained in the following.

We implemented a Webots gait demonstration with the following parameters: locomotion frequency $f=\SI{3.5}{Hz}$, desired step length of \SI{12}{cm}, fore and hind touch down angle around $\SI{2.85}{rad}$. To maximize the available leg retraction, a foot lift-up height of \SI{4}{cm} was set. The virtual, commanded duty factor was \num{0.49}, the observed effective duty factor \num{0.52} and \num{0.58} for fore and hindlimb, respectively. Remaining control parameters were optimized with a PSO framework~\citep{kennedy_pso_1995} to maximize covered distance over a \SI{15}{s} time interval, after a \SI{5}{s} time window to reach steady-state. Snapshots of the Webots Oncilla robot trotting are shown in~\cref{fig:webots-trot}, the corresponding video link is given in Table S1 of the supplementary material.

\begin{figure}[h!]
\begin{center}
\includegraphics[width=.9\textwidth]{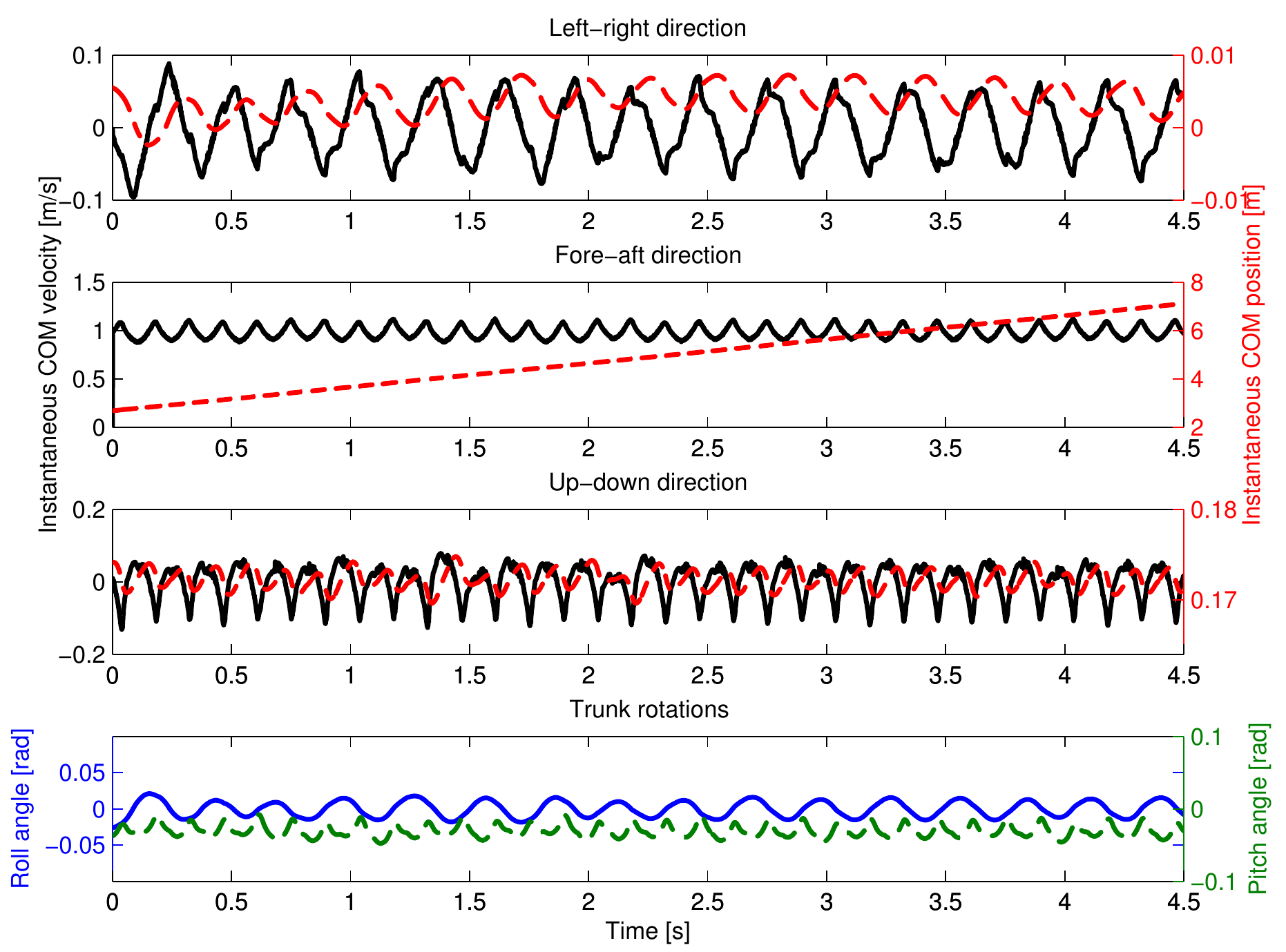}
\end{center}
\caption{
Experiment in Webots. 
Results are plotted same style as hardware results in \cref{fig:COM-speed}. 
The simulated robot's center of mass (COM) position and velocity are shown over time, for a cycle frequency of \SI{3.5}{Hz}. 
Shown are instantaneous COM position (dashed red) and velocity (full, black line) components sorted by their left-right, fore-aft, and up-down components. 
The robot reached an average trotting speed of \SI{0.98}{\mps}, at peak speeds of \SI{1.12}{\mps}. 
Vertical COM displacement was \SI{\pm 2}{mm}, at an average hip height of \SI{0.17}{m}. 
Average roll angle was \SI{\pm 0.015}{rad}. 
The average pitch angle was \num{-0.03} \SI{\pm 0.015}{rad}.}
\label{fig:COM-speed-webots}
\end{figure}

From results we recognize the relatively large speed gap for forward locomotion between hardware and Webots simulation at equal locomotion frequencies (\cref{fig:COM-speed,fig:COM-speed-webots}, \SI{3.5}{Hz}). 
The hardware robot reached a maximum forward speed of \SI{0.63}{m/s}, with \SI{0.09}{m} effective stride length.
The Webots model reached a forward speed of \SI{0.98}{m/s}, with \SI{0.14}{m} effective stride length.
The Webots robot's COM oscillated $\pm \SI{2}{mm}$ vertically. 
The simulated robot showed roll angle oscillating around $\SI{0}{rad}\pm\SI{0.015}{rad}$, and pitch angles oscillating around $\SI{-0.03}{rad}\pm\SI{0.015}{rad}$.

Further analysis of the foot locus shows the optimization framework increased the model robot's speed by making precise use of the robot's toe segment (Fig.\,S1).
This increased effective stride by \SI{15}{\%}, from \SI{0.12}{m} to \SI{0.14}{m}.
In hardware, we observed a decrease of effective stride length, compared to the commended stride length.
At frequencies above \SI{3.5}{Hz} the hardware robot started slipping, and the effective stride length reduced to \SI{0.09}{m}. 

The gap between simulation and real-hardware speed makes it hard to one-to-one transfer control parameters for open-loop gaits. However, in parallel to this work, Oncilla robot's Webots model was successfully applied to prototype a closed loop controller~\citep{mos-thesis,ajallooeian2013modular}. The transfer of the closed-loop controller onto the hardware was done with minimal effort, and showed a better performance matching compared to the open-loop parameter transfer shown here.

 \section{Discussion}\label{sec:discussion}
Oncilla robot is a small, light-weight, quadruped legged robot with compliant, spring-loaded pantograph legs and three active degrees of freedom per leg. 
The robot trots open-loop indoors and outdoors, with forward speeds up to \SI{0.63}{\mps} ($\mathrm{Fr}=\num{0.25}$).
It climbs up to \ang{10} slopes at a speed of \SI{0.25}{\mps} backwards, and \ang{4} slopes at a forward speed of \SI{0.15}{\mps}. 
All hardware experiments shown are with a single controller, for trotting, in open-loop.
The robot descends slopes of \ang{10} at a speed of \SI{0.42}{\mps} backwards, and \SI{0.40}{\mps} forwards.
Oncilla robot's adduction/abduction (AA) joints permit fast turning maneuvers, with smaller turning radii compared to turning maneuvers utilizing only leg angle and leg length joints.
Utilizing AA joints enables turning on the spot, with \SI{10}{\second} for a full turn. Furthermore, AA turning allows the robot to turn while mostly maintaining the commanded speed, i.e.\,the robot would lose as little as \SI{20}{\%} speed while turning on a radius of \SI{0.46}{m}.
Oncilla robot showed best cost of transport (COT) at a locomotion speed of \SI{0.63}{\mps}, with $\mathrm{COT} = \SI{3.2}{J/(Nm)}$.
A set of sensors is available to monitor the robot's posture, ground contacts, and trunk-leg reaction forces for feedback control.
A high bandwidth, real-time (\SI{500}{Hz}) network allows communication between the robot's electronic boards.
The robot's main controller runs at \SI{200}{Hz} in real time.
The robot's software architecture allows to transparently control the robot either onboard, or remotely, with multiple programming interface options. 
It further allows to control hardware and Webots simulation with the same controller architecture.
Oncilla robot's multibody dynamics simulation in Webots can be used for fast prototyping controller architectures.

Oncilla robot incorporates several features of Cheetah-cub robot~\citep{sproewitz_ijrr}. We implemented additional turning capabilities through Oncilla robot's AA DOF, with a set of RC servo motors rotating the robot's leg in an axis parallel to the robot's rolling axis. Due to the nature of serial joints, the entire leg structure and its actuators are rotated by the AA joint. This worked satisfactorily, for the shown examples. In case of harsh perturbations or for agile side stepping motions, a more powerful AA joint will become necessary.

Oncilla robot's spring-loaded, pantograph legs performed well for level locomotion indoors and outdoors, slope trotting upwards and downwards, turning during locomotion, and turning on the spot. The leg's intrinsic compliance resolves overdetermined kinematic loops, and all experiments could be performed open-loop. Closed-loop control with Oncilla robot is possible for unstructured terrain~\citep{mos-thesis,ajallooeian2013central}. The current design has a limitation which we are planning to resolve in the future: the leg extends only by releasing its charged extensor springs. For maneuvers demanding higher leg power, such as jumping or leaping, an additional actuator, or a redesign of the robot's flexor actuator into a flexor-extensor actuator will become necessary. 

One serious problem with RC servo actuated robots is actuator saturation. 
At combined loads of high speed and torque, these actuators heat up above their thermal dissipation capabilities and required regular cool-down stops as with Cheetah-cub robot~\citep{sproewitz_ijrr}.
Hence, Oncilla robot's LL and LA actuators were designed for a low cost of transport, and for sufficient thermal capacities through larger, brushless motors, and optimized gearbox ratios. 
This lowers the robot's relative power intake, and allows to run it without exceeding hand-warm motor temperatures, at high load and up to \SI{4}{Hz} cycle frequency. The relatively low COT allows the robot to run tether-free up to \SI{30}{min} with a \SI{4500}{mAh} lithium polymer battery.
In open-loop mode, good speed can be achieved, i.e. \SI{0.63}{\mps} ($\mathrm{Fr}=\num{0.25}$). Optimizing locomotion control parameters, and introducing feedback will likely yield higher robot speed.
We expect that continuous progress with off-the-shelf components will further reduce robot weight and increase robot performance: 
more compact, powerful, brushless motors of different form factors are becoming available~\citep{de_penn_2015}, together with compact, high-power, four-quadrant brushless motor controllers~\citep{vedder_vesc_nodate}.

Oncilla robot's best measured hardware cost of transport of \SI{3.2}{J/(Nm)} is \SI{54}{\%} lower than Cheetah-cub robot's COT of \SI{6.9}{J/(Nm)}. \cite{tucker_energetic_1970} shows that the metabolic cost of transport of animals decreases with increasing body weight. 
This heuristically found relationship predicts a COT of \SI{1.0}{J/(Nm)} for an animal of the weight of Oncilla robot ($m=\SI{5}{kg}$).
The currently most energy efficient, fully active quadruped robot is the \SI{33}{kg} MIT Cheetah, featuring a COT of \SI{0.5}{J/(Nm)} at fast, \SI{6}{m/s} trotting~\citep{hyun_high_2014}. 
It benefits from custom made brushless actuators, and energy recuperation capabilities.

Oncilla robot trotted slopes with inclination angles up to \ang{10}, faster and with less slippage while going backwards. As nothing else changed,  the speed discrepancy must have been caused by leg asymmetry, between forward and backward orientated feet, and pantograph leg joints. More research is required to understand the effects of underactuated, segmented and spring-loaded legs.

By open-sourcing Oncilla robot's mechanical, electrical, software, and simulation blueprints, we aim towards an easily accessible platform for research and education. We hope this project will help spawning improved robots, and allow the field to grow and extend rapidly. All source files are online accessible through permanent repositories, links are provided in~\cref{tab:sources}. This project was shared between multiple universities in the framework of the EU-FP7 AMARSi project, and five Oncilla robots have been distributed among project partners. 

Recently, fluidic and fluidic-hybrid actuation became an interesting choice for articulated robots research~\citep{whitney_hybrid_2016}. Pneumatically actuated quadruped robots like Pneupard~\citep{rosendo_quadrupedal_2014} can utilize the actuator's intrinsic compliance, and high frequencies. This keeps robot weight and complexity low. At the same time, large torques, angular amplitudes, and locomotion frequencies even above \SI{7}{Hz} are feasible~\citep{narioka2012development}. This present an interesting option for future electric-pneumatic hybrid legged robots.

Soon, legged robots will need to keep up with the claim to perform better than wheeled or tracked vehicles in cluttered environments, and over rough terrain. Boston Dynamics LittleDog robot showed great results as the common robot platform in an earlier DARPA challenge~\citep{righetti_optimal_2013}. During the tasks of climbing over obstacles, LittleDog robot benefited from its three DOF leg design, and a very large motion joint range. Other legged robots like StarlETH robot~\citep{hutter_efficient_2013} or HyQ~\citep{semini_design_2011} also implemented \num{3} DOF legs with large joint ranges. Oncilla robot's AA joint allows the robot to turn on the spot, and trot efficiently during turning. For tasks in rougher terrain, or to stabilize the robot from sudden sideways perturbations a faster and stronger AA joint with a larger motion range will become necessary. 

We found Oncilla robot's small form factor very helpful. Production and component costs are relatively low, and maintenance is easy. Due to its low weight the robot can safely be handled by a single user, and without gantry installations.

Finally, Oncilla robot provides sensor data, of joint and motor encoders, from gyroscopes, forces sensors, and ground contact sensing based on leg spring deflection. This allows research of closed-loop control~\citep{mos-thesis,ajallooeian2013modular,ajallooeian2013central}, and testing bioinspired locomotion controllers. It also makes Oncilla robot a compact and relatively low cost mobile gait analysis tool. Oncilla robot has remaining load capacities for extra sensors like stereoscopic cameras, or distally mounted leg-force sensors. The robot's communication interface and software architecture allows to include such sensors transparently.

\section{Conclusion}
\label{sec:conclusion}
We presented a novel compliant quadruped robot, its software and control framework, and its Webots simulation. We characterize Oncilla robot trotting on level ground, climbing and descending slopes, and during turning. Locomotion speed and direction were set by an open-loop controller in all experiments, without the need to track the robot's posture, internal leg constraints, or considering foot slippage. 
By outsourcing control tasks into mechanics, Oncilla robot can locomote with low control effort, i.e.~open-loop and low control frequencies, over different in- and outdoor substrates. 
Oncilla robot reached a maximum forward speed of \SI{0.63}{\mps} at a transport of COT$=\SI{3.2}{J/(Nm)}$, at \SI{5}{kg} weight.
The robot can turn on the spot, or with a very small turning radius, depending on the applied controller and the utilized robot DOF. 
The presented example locomotion controller implements gait features in a modular manner, through a central pattern generator. 
Furthermore, Oncilla robot's open software architecture allows testing other control approaches. 
For future research with closed loop control, sensors are incorporated and ready for use, i.e.\,reaction force sensors, gyroscopic sensing, leg-ground detection, and absolute joint position sensors.
Oncilla robot's mechanical and electrical blueprints, firmware, simulation, and software architecture are open-source available under GPLv3~\citep{gplv3} for hardware and firmware, and LGPLv3~\citep{lgplv3} for driver, simulation and software architecture. Links are available in~\cref{tab:sources}. The supplementary document for this manuscript includes a more detailed description of Oncilla robot's electronic hardware components, a table with links to videos, schematic figures and kinematic variables of the robot's leg morphology, and details to the knee and hip motor optimization applied.

\section*{Links to videos, via YouTube}
\begin{enumerate}
\item Oncilla robot trots backwards and forwards \url{https://youtu.be/38pX1FBRlEA}
\item Oncilla robot trotting up a \SI{4}{\degree} slope \url{https://youtu.be/c7wudgzZNkc} 
\item Oncilla robot turning on the spot, real-time \url{https://youtu.be/TH8AB1mdSoY}
\item Oncilla robot in outdoor environment \url{https://youtu.be/A20KLlwuwTg}
\item Webots simulation of Oncilla robot trotting forward \url{https://youtu.be/0eAhhNvKjGM}
\end{enumerate}


\section*{Funding}
The research leading to these results has received funding from the European Community's Seventh Framework Programme (FP7/2007-2013 Challenge Cognitive Systems, Interaction, Robotics; grant agreement number 248311 (AMARSi)) and from the Swiss National Science Foundation through the National Centre of Competence in Research Robotics.
The work of Alexandre Tuleu was supported by the scholarship SFRH/BD/51451/2011 from Fundacao para e Cienca e a Tecnologia.
 \section*{Acknowledgments}
We thank Jesse van den Kieboom, who developed code, and provided infrastructure for the PSO-based optimization framework. We thank A.\,Crespi, A.\,Guignard, M.\,Heynick, F.\,Longchamp, and Y.\,Bourquin for technical support.

\bibliographystyle{frontiersinSCNS_ENG_HUMS} 
\bibliography{Biblio}

\section*{Figures}

 \begin{table}[h]
\caption{
Physical parameters of Oncilla robot, leg masses without adduction/abduction (AA) motors. Abbreviations: front hip (FH), fore-aft (FA) plane, lateral plane (LAT), vertical plane (VERT), leg length (LL), gear box (GB). Center of mass (COM) position relative to the robot's geometric center, based on the computer aided design. }
\small \centering
\begin{tabular}{ll}
Parameter \si{[unit]}														& Value \\
\hline
Overall length/height/width \si{[m]}								   & 0.40/0.31/0.25 \\
Leg length min/standing/max  \si{[m]}									& 0.11/\num{0.16}/\num{0.18}\\
Hip spacing (FA)  \si{[m]}													& 0.23 \\
Hip spacing AA axes lateral  \si{[m]}										& 0.14 \\
Leg angle range (FA)  \si{[\degree]}									& \num{\pm 34} \\
Leg angle range AA \si{[\degree]}									& \num{\pm 8}    \\
Robot mass without/with battery \& cables \si{[kg]}					& \num{4.5}/5.1 \\
COM position vertical \si{[m]}											& \num{-0.036} \\
COM position FA, from FH  \si{[m]}										& \num{-0.030}     \\
Leg mass without/with actuators \si{[kg]}							   & 0.22/\num{0.6}\\
Hip/LA torque at \SI{6}{A} at GB ratio $n=84$ \si{[Nm]}		   & 7.1 \\
LL torque at \SI{6}{A} at GB ratio $n=56$ \si{[Nm]}            & 4.7 \\
Hip (AA) stall torque \si{[Nm]}											& 2.9
\end{tabular}
\label{tab:physical-parameters}
\end{table}

\begin{table}
\caption{
Hardware experiment. 
Results for locomotion on inclined surfaces of different slopes while ascending or descending. 
Locomotion on up slopes was more successful when trotting backwards. 
Abbreviations: commanded velocity ($\mathrm{v}_\mathrm{cmd}$), average velocity ($\mathrm{v}_\mathrm{avg}$), inclination (Inc), excessive slippage (ES).}
\label{tab:inclined}
\centering

\begin{tabular}{rrr}
Inc & $\mathrm{v}_\mathrm{cmd}$ & $\mathrm{v}_\mathrm{avg}$ \\ 
	$[\si{\degree}]$        & $[\si{\mps}]$          & $[\si{\mps}]$     \\
\hline
+4		& 0.4			& 0.15	\\
+4    & -0.4		& -0.33	\\
+7    & 0.4			& ES		\\
+7    & -0.4		& -0.27	\\
+10   & 0.4			& ES		\\
+10   & -0.4		& -0.25	\\
-10	& 0.4			& 0.40	\\
-10	& -0.4		& -0.42	\\
\end{tabular}
\end{table}

\begin{table}
\caption{
Results for turning with two different strategies: 
{\inlinelist{\protect\item  Amplifying the movement of the shoulder/hip abduction/adduction degree of freedom (AA amp). \protect\item Asymmetrically shortening the stride length (ASL) on one side of the robot.}} 
Scale factor (SF): when \num{0}, stride length on both sides are the same, when \num{0.5}, stride length on one side is zero, and when \num{1}, stride length on one side is reversed. Abbreviations: commanded velocity ($\mathrm{v}_\mathrm{cmd}$), amplitude of AA movement ($\mathrm{A}_\mathrm{AA}$), turning radius (r), time required for a full \SI{360}{\degree} turn ($\mathrm{t}_\mathrm{FT}$), and average resulting velocity ($\mathrm{v}_\mathrm{avg}$) are shown.}

\label{tab:turning}
\centering

\begin{tabular}{lllllrl}
	Strategy &   $\mathrm{v}_\mathrm{cmd}$ & SF & $\mathrm{A}_\mathrm{AA}$ & r & $t_\mathrm{FT}$ & $\mathrm{v}_\mathrm{avg}$ \\ 
				& [\si{\mps}] & [] & [\si{rad}] & [\si{m}] & [\si{s}] & [\si{\mps}] \\
  \hline
  \multirow{3}{*}{AA ampl.} 
 	  & 0.4 & 0.0 & 0.05 & 0.46 & 9 & 0.32 \\ 
 	  & 0.2 & 0.0 & 0.05 & 0.23 & 10 & 0.14 \\ 
 	  & 0.0 & 0.0 & 0.05 & $\approx0$ & 10 & $\approx0$ \\ 
  \hline
  \multirow{4}{*}{ASL} 
 	  & 0.4 & 0.3 & 0.00 & 0.50 & 17 & 0.19 \\ 
     & 0.4 & 0.4 & 0.00 & 0.32 & 14 & 0.15 \\ 
 	  & 0.4 & 0.5 & 0.00 & 0.19 & 9 & 0.13 \\ 
 	  & 0.4 & 1.0 & 0.00 & 0.03 & 7 & 0.03 \\ 

\end{tabular}
\end{table}

\begin{figure}[h!]
\begin{minipage}[b]{.5\linewidth}
\centering\includegraphics[width=.95\textwidth]{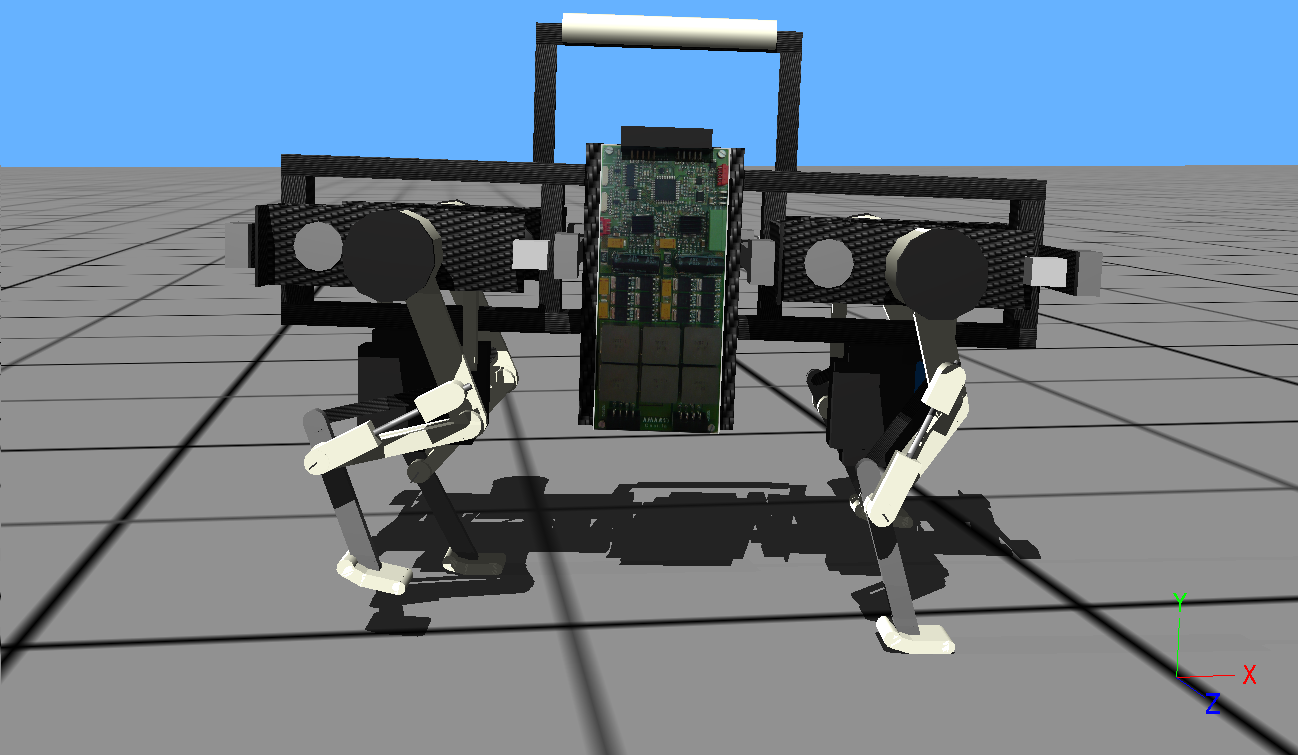}
\end{minipage}
\begin{minipage}[b]{.5\linewidth}
\centering\includegraphics[width=.95\textwidth]{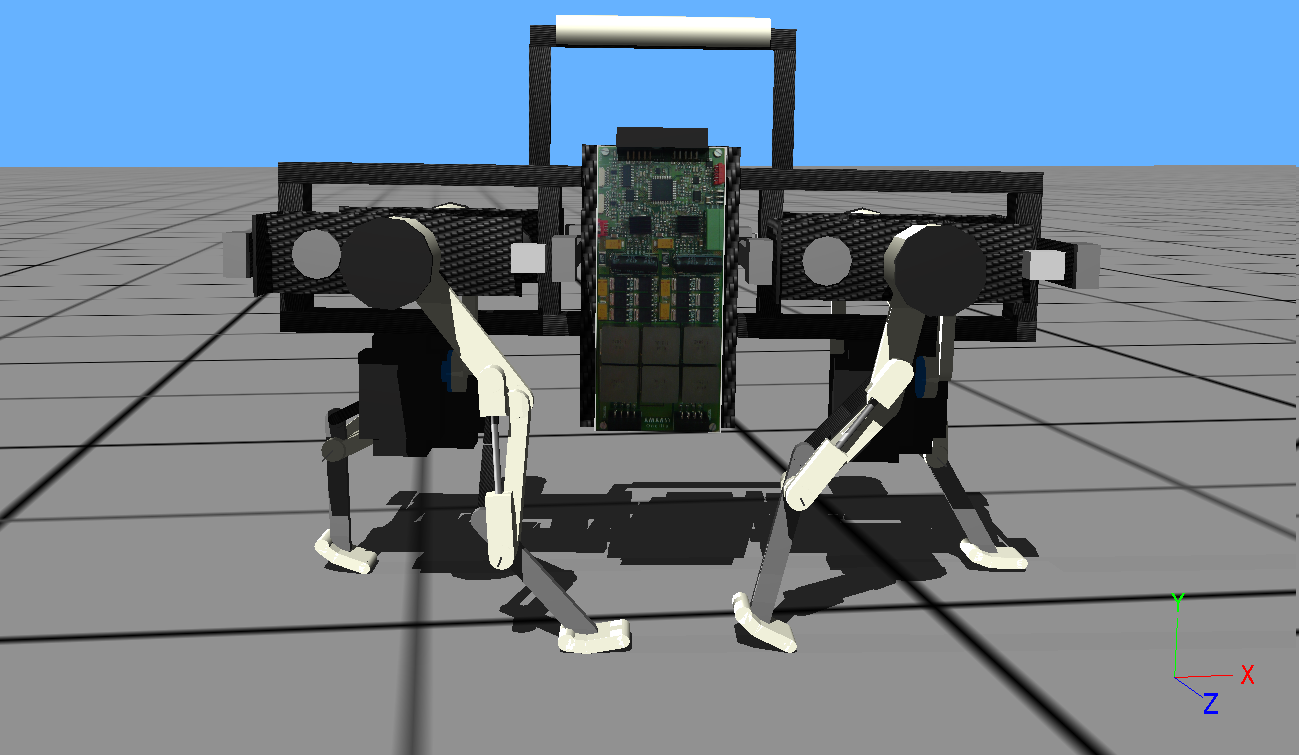}
\end{minipage}
\caption{
Snapshots show Oncilla robot simulated in Webots, with the robot's heading to the right.
The robot trots with an average speed of \SI{0.98}{\mps}, at locomotion frequency \SI{3.5}{Hz}. 
The commanded duty factor is \num{0.49}, the observed duty factor is \num{0.52} and \num{0.58} for front and hind legs, respectively. 
The commanded step length was \SI{0.12}{m}, the observed step length \SI{0.14}{m}. 
For this run, the robot accelerated from its still standing position, without controlled transition. 
The corresponding video shows the acceleration as short trunk pitching (link in Table S1).
}
\label{fig:webots-trot}
\end{figure}

 \begin{table}[]
\centering
\caption{Link list Oncilla robot permanent open-source repositories. Unless specified otherwise (i.e.\,'apt' or 'wget'), users can Git clone repositories. All repositories can be open-source utilized in accordance to the specified license (LGPL-3 or GPL-3).}\label{tab:sources}
\centering
\tiny
\begin{tabularx}{\textwidth}{llX}
Name                    & License & Format / description / on-line location                                                             \\
\hline
Oncilla Hardware         & GPL-3   & SolidWorks 2014, Altium designer, PDF summary                                                      \\
Blueprint &         & Mechanical and Electronics Blueprints of the Oncilla robot                                        \\
                        &         & \url{https://c4science.ch/source/oncilla-hardware/}                                  \\
sbcp-uc                 & GPL-3   & C sources, MPLAB-X project                                                                         \\
                        &         & SBCP protocol slave implementation for dsPIC33FJ DSP processor family                              \\
                        &         & \url{https://c4science.ch/source/sbcp-uc/}                                          \\
Oncilla SBCP  & GPL-3   & C sources, MPLAB-X project                                                                         \\
Master Firmware                       &         & SBCP Master Communication board firmware                                                           \\
                        &         & \url{https://c4science.ch/source/oncilla-sbcp-master-fw/}			       \\
Oncilla Power Management         & GPL-3   & C sources, MPLAB-X project                                                                         \\
Board Firmware                        &         & Power board firmware                                                                               \\
                        &         & \url{https://c4science.ch/source/oncilla-power-fw/} 				       \\
Oncilla Motordriver   & GPL-3   & C sources, MPLAB-X project                                                                         \\
Firmware                        &         & Motordriver board firmware                                                                         \\
                        &         & \url{https://c4science.ch/source/oncilla-motordriver-fw/} 			       \\
SBCP                    & LGPL-3  & C sources, CMake project                                                                           \\
Low-Level Driver                        &         & Low-level SBCP, user-space, device agnostic SBCP drivers, based on libftdi                         \\
                        &         & \url{https://c4science.ch/source/sbcpd/} 					       \\
SBCP                  & LGPL-3  & C++ sources, CMake project                                                                         \\
High-Level Driver                        &         & Object oriented, high-level, device specific SBCP drivers, Bus management Command Line Utilities   \\
                        &         & \url{https://c4science.ch/source/libsbcp/} 					       \\
liboncilla              & LGPL-3  & C++ sources, CMake project                                                                         \\
                        &         & Level-0 RCI interface, depends on librci \citep{nordmann2012software}                              \\
                        &         & \url{https://c4science.ch/source/liboncilla/} 				       \\
liboncilla Hardware           & LGPL-3  & C++ sources, CMake project                                                                         \\
Implementation                        &         & Level-0 Back-end implementation for the Oncilla hardware                                            \\
                        &         & \url{https://c4science.ch/source/liboncilla-hw/} 				       \\
liboncilla Webots        & LGPL-3  & C++ sources, CMake project                                                                         \\
Interface                        &         & Level-0 Back-end implementation for the Oncilla Webots based simulator                              \\
                        &         & \url{https://c4science.ch/source/liboncilla-webots/} 				       \\
Oncilla Simulation              & LGPL-3  & Python sources                                                                                     \\
                        &         & Helper program to setup Oncilla simulation environment                                           \\
                        &         & \url{https://c4science.ch/source/oncillasim/} 				       \\
CoR-Lab APT repository  & n.a.    & Apt repository                                                                                     \\
                        &         & Apt (Debian/Ubuntu) repository for liboncilla dependencies (precise and trusty supported)         \\
                        &         & \url{http://packages.cor-lab.de} 								       \\
Biorob APT repository   & n.a.    & Apt repository                                                                                     \\
                        &         & Apt (Debian/Ubuntu) repository for liboncilla and liboncilla-webots (precise and trusty supported) \\
                        &         & \url{https://ponyo.epfl.ch/packages} 
\end{tabularx}
\end{table}

\end{document}